\documentclass[lettersize,journal]{IEEEtran}
\usepackage{amsmath,amsfonts}
\usepackage{array}
\usepackage[caption=false,font=normalsize,labelfont=sf,textfont=sf]{subfig}
\usepackage{textcomp}
\usepackage{stfloats}
\usepackage{url}
\usepackage{verbatim}
\usepackage{graphicx}
\usepackage{cite} 
\usepackage{xcolor} 
\usepackage{colortbl}
\usepackage{float} 
\usepackage{tabularx}  
\usepackage{wrapfig}  
\usepackage{multirow}
\usepackage{makecell} 
\usepackage{changepage}  
\usepackage{pifont}
\usepackage{bm}
\usepackage{dsfont} 
\usepackage{tikz} 
\usepackage{makecell} 
\usepackage{booktabs}  
\usepackage{bbding} 
\usetikzlibrary{arrows.meta, bending, calc,arrows}

\definecolor{myviolet}{RGB}{168,77,153} 
\definecolor{myblue}{RGB}{60,118,185} 
\definecolor{cvprblue}{rgb}{0.21,0.49,0.74}
\definecolor {mygreen}{RGB}{70, 145, 80} 
\newcommand{\xl}[1]{\textcolor{black}{#1}}

\def\BibTeX{{\rm B\kern-.05em{\sc i\kern-.025em b}\kern-.08em
    T\kern-.1667em\lower.7ex\hbox{E}\kern-.125emX}}
\usepackage{balance}

\usepackage{amsmath,amssymb,amsfonts}
\usepackage{amsthm}
\usepackage{algorithm}
\usepackage{algorithmicx}
\usepackage{algpseudocode}
\usepackage{array}
\usepackage{arydshln}
\usepackage{balance}
\usepackage{bbding}
\usepackage{bm}
\usepackage{booktabs}
\usepackage{cite}
\usepackage[table]{xcolor}
\usepackage{colortbl}
\usepackage{dsfont}
\usepackage{float}
\usepackage{graphicx}
\usepackage{listings}
\usepackage{makecell}
\usepackage{mathrsfs}
\usepackage{microtype}
\usepackage{multirow}
\usepackage{nccmath}
\usepackage{pifont}
\usepackage[caption=false,font=normalsize,labelfont=sf,textfont=sf]{subfig}
\usepackage{textcomp}
\usepackage{url}
\usepackage{verbatim}
\usepackage{wrapfig}
\usepackage[colorlinks,linkcolor=red,anchorcolor=blue,citecolor=green]{hyperref}
\usepackage{makecell}

\def\BibTeX{{\rm B\kern-.05em{\sc i\kern-.025em b}\kern-.08em
    T\kern-.1667em\lower.7ex\hbox{E}\kern-.125emX}}

\let\citep\cite
\let\citet\cite
\newcommand{\sssection}[1]{\noindent\textbf{#1}}
\makeatletter

\newcommand{\Rmnum}[1]{\expandafter\@slowromancap\romannumeral #1@}
\newcommand{\thickhline}{%
    \noalign {\ifnum 0=`}\fi \hrule height 1pt
    \futurelet \reserved@a \@xhline
}
\makeatother

\definecolor{cred}{HTML}{FF6B6B}
\definecolor{cyellow}{HTML}{FEC260}
\definecolor{cgreen}{HTML}{70AD47}
\definecolor{cblue}{HTML}{4D96FF}
\definecolor{cpurple}{HTML}{2A0944}
\definecolor{ggray}{RGB}{127,127,127}
\definecolor{aliceblue}{rgb}{0.94, 0.97, 1.0}
\definecolor{mygray}{gray}{.9}
\definecolor{mydarkblue}{rgb}{0,0.08,0.45}
\definecolor{chatgptgreen}{RGB}{16,163,127}
\definecolor{cvprblue}{rgb}{0.21,0.49,0.74}

\theoremstyle{plain}

\theoremstyle{definition}

\newcommand{\pub}[1]{\color{gray}{\tiny{[{#1}]}}}

\begin{document}

\title{Hierarchical Relation-augmented Representation Generalization for Few-shot Action Recognition}

\author{Hongyu~Qu, Ling~Xing, Jiachao~Zhang,~\IEEEmembership{Member, IEEE}, Rui~Yan, Yazhou~Yao,~\IEEEmembership{Member, IEEE}, and Xiangbo~Shu,~\IEEEmembership{Senior Member, IEEE}
\thanks{\textit{H. Qu, L. Xing, R. Yan, Y. Yao, and X. Shu are with the School of Computer Science and Engineering, Nanjing University of Science and Technology, Nanjing 210094, China. E-mail: \{quhongyu, lingxing, ruiyan, yazhou.yao, shuxb\}@njust.edu.cn. (Corresponding author: Xiangbo Shu.)}}
\thanks{\textit{J. Zhang is with the Artificial Intelligence Industrial Technology Research Institute, Nanjing Institute of Technology, Nanjing 211167, China (e-mail: zhangjc07@foxmail.com).}}
}

\markboth{The Submission of IEEE Transactions on Image Processing, 2026}%
{Qu \MakeLowercase{\textit{et al.}}: Hierarchical Relation-augmented Representation Generalization for Few-shot Action Recognition}

\maketitle

\begin{abstract}
Few-shot action recognition (FSAR) aims to recognize novel action categories with limited exemplars. Existing methods typically learn frame-level representations independently for each video by designing various inter-frame temporal modeling strategies. However, these methods suffer from ``cognitive limitation” regarding information isolation, treating each video and task as an independent entity.
This information isolation contradicts human cognitive mechanisms, which dynamically associate the current observation with concurrent examples to distill common temporal patterns (inter-video correlation)  and recall relevant motion priors from long-term memory (inter-task knowledge transfer).
In light of this, we propose HR$^{2}$G-shot, a Hierarchical Relation-augmented Representation Generalization framework for FSAR, which mimics the human cognitive process by unifying three types of relation modeling (inter-frame, inter-video, and inter-task) to learn task-specific temporal patterns from a holistic view. 
In addition to conducting inter-frame temporal interactions, we further devise two cognitively-inspired components to respectively explore inter-video and inter-task relationships: \textbf{i)} Inter-video Semantic Correlation (ISC) performs cross-video frame-level interactions in a fine-grained manner, thereby capturing task-specific query features and learning intra- and inter-class temporal correlations among support features; \textbf{ii)} Inter-task Knowledge Transfer (IKT) retrieves and aggregates relevant temporal knowledge from the bank, which stores diverse temporal patterns from historical tasks. Extensive experiments on five benchmarks show that HR$^{2}$G-shot outperforms current top-leading FSAR methods.
\end{abstract}


\section{Introduction}
During the last few years, action recognition~\citep{feichtenhofer2019slowfast,xing2023svformer,huang2023semantic,cao2024eventcrab,li2024ftmomamba,xu2026attack,jiao2026rethinking,li2026star,zhou2025zero} has witnessed remarkable progress with the advances in deep learning.
However, their strong performance heavily relies on a large amount of labeled training examples, which restricts the model's scalability and generalizability in data scarcity scenarios.
In contrast, human beings can easily learn new visual concepts with only a few supervisions.
To enable the machine to acquire such ability, many studies have shifted their attention towards few-shot action recognition (FSAR)~\citep{zhu2018compound,cao2020few,wang2024hyrsm++,gao20252}, which aims to learn novel (unseen) action classes using only a few annotated video samples after training on a set of base (seen) classes with abundant samples.

Current top-leading FSAR solutions typically adopt the metric-based meta-learning paradigm, wherein the model first maps the query videos and support videos into a discriminative feature space, and then performs query-support video matching based on pre-defined or learned distance metrics. 
To achieve this, a prevalent number of subsequent efforts delve into network designs for \textbf{\textit{temporal representation learning}} by, \emph{e.g.}, temporal attention operations~\citep{wang2022hybrid}, elaborated spatio-temporal~\citep{thatipelli2022spatio} or temporal-channel interaction~\citep{boosting}, multi-layer feature fusion~\citep{xing2023revisiting}, and auxiliary modal information (\emph{e.g.}, depth~\citep{fu2020depth} and motion~\citep{wang2023molo}). 
Apart from investigating \textit{temporal representation learning}, to perform proper temporal feature alignment when comparing query and support videos, recent approaches devise \textbf{\textit{various temporal matching strategies}}, \emph{e.g.}, frame-level feature matching~\citep{cao2020few,wang2024hyrsm++}, tuple-level feature matching~\citep{perrett2021temporal}, frame-to-segment feature matching~\citep{wu2022motion}, or multi-level feature matching~\citep{zheng2022few}.

\begin{figure*}[!t]
    \centering
        \includegraphics[width=0.96\textwidth]{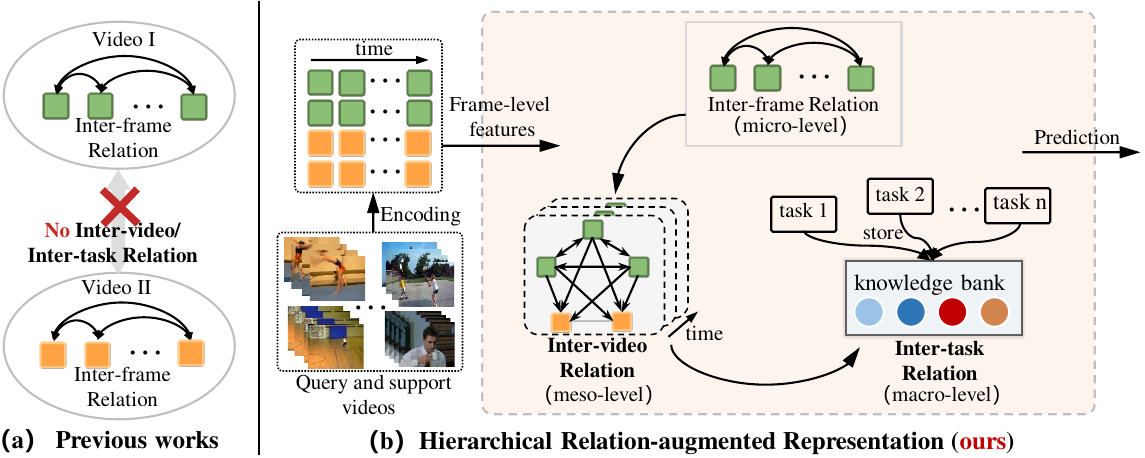}
        \vspace{-8pt}
		\caption{Our main idea. (a): Previous FSAR works only rely on inter-frame relation modeling to learn video representations, ignoring the relations between videos and tasks. (b): In contrast, we unify three types of relation modeling (\emph{i.e.}, inter-frame, inter-video, inter-task) under one single framework, so as to capture task-specific temporal cues.}
		\label{fig::motivation}	
        \vspace{-4.3mm}
	\end{figure*}

Despite their performance, we argue that these methods suffer from ``cognitive limitation" regarding information isolation (Fig.~\ref{fig::motivation}a).
\textbf{First}, existing methods directly learn meaningful video representations from one single task, \textbf{\textit{lacking an explicit mechanism for longitudinal knowledge accumulation (inter-task relations)}}. Consequently, they fail to distill universal temporal patterns that are shared across diverse actions. Actually, novel actions are often variations or compositions of previously learned motion primitives. Ignoring this inter-task relatedness prevents the model from ``warm-starting" with prior experience, causing sub-optimal transferability and generalization on new tasks with only a few samples.

\textbf{Second}, they independently learn frame-level representations for each video, \textbf{\textit{neglecting the collective semantics (inter-video relations) in the current task}}.
Without cross-referencing multiple examples, models struggle to distinguish task-relevant motion patterns from instance-specific noise (\emph{e.g.}, complex backgrounds or camera motion).
Exploring inter-video relationships allows the model to reach a ``consensus" on the common action essence (\emph{e.g.}, action styles and speed)  while filtering out non-shared discrepancies. Although some preliminary attempts~\citep{boosting,wang2022hybrid} have been made towards this, they capture sample-level correlations at the video level (global feature for each video), ignoring fine-grained temporal details in videos. Thus, these methods fail to capture how specific motion segments align across different samples.  

From a cognitive perspective, these limitations stem from a fundamental deviation from human intelligence. Human capability is rooted in a \textbf{holistic cognitive mechanism}~\citep{lake2015human}: when humans learn a new action, they do not view events in isolation. Instead, they dynamically associate the current observation with concurrent examples to distill common patterns (contextualization)~\citep{gentner1983structure} and recall relevant motion priors from long-term memory (knowledge transfer)~\citep{zacks2007event}. 
Without such a mechanism, existing models struggle to capture the intrinsic structure of actions from a few examples, creating a significant gap in sample efficiency compared to human learners. In light of the above, our goal is to \textbf{\textit{shift the paradigm from isolated instance learning to hierarchical relational learning (i.e., inter-frame, inter-video, inter-task) under one single framework}}, so as to yield task-augmented temporal features for each video via transferring knowledge at both video-level and task-level (as shown in Fig.~\ref{fig::motivation}b).

In this vein, we develop a novel hierarchical framework for FSAR, namely HR$^{2}$G-shot, which mimics the human cognitive process by progressively modeling temporal cues across three perspectives: micro-level (inter-frame), meso-level (inter-video), and macro-level (inter-task). 
By unifying these hierarchies, we enable the model to not only parse internal motion details (\emph{i.e.}, inter-frame temporal interactions) but also make use of external relationships for holistic understanding. To achieve this, we propose two cognitively-inspired modules:
\textbf{i)} \textbf{Inter-video Semantic Correlation (ISC)} aims to precisely capture meso-level contextualization within each task. Specifically, we perform cross-video frame-level interactions in a fine-grained manner to gain task-specific temporal features and reduce considerable computational costs compared with inter-video dense interaction (see Table~\ref{tab::fine-attention}). Moreover, to respect the difference between support and query videos, we customize the masked interaction strategy for support-support and query-support relations (\emph{i.e.}, support and query videos only aggregate semantic information from support videos). This asymmetric design ensures that query videos actively aggregate semantic cues from support anchors, while support videos reinforce their intra-class compactness, thereby yielding robust task-specific representations.
 \textbf{ii)} \textbf{Inter-task Knowledge Transfer (IKT)} is devised to realize macro-level knowledge transfer between tasks by exploring inter-task relationships. Concretely, we develop a temporal knowledge bank that serves as the model's long-term memory, storing diverse temporal patterns from historical tasks. Then we make use of temporal prototypes, which summarize frame-level representations of each video, to retrieve and aggregate temporal knowledge related to new tasks from the bank. Integrated with such knowledge, our method yields more meaningful video representations, thus quickly adapting to new tasks with only a few samples. We conduct extensive experiments on five gold-standard datasets, and the results demonstrate that our HR$^{2}$G-shot outperforms existing top-leading FSAR methods by a large margin.

Overall, the main contributions of this work are threefold:
\begin{itemize}
   \item We provide a hierarchical relation-augmented framework for FSAR, which mimics the human cognitive process by progressively modeling temporal cues across three perspectives: micro-level (inter-frame), meso-level (inter-video), and macro-level (inter-task). 
   \item \textbf{To capture collective semantics within each task}, we propose Inter-video Semantic Correlation module that conducts cross-video frame-level interactions in a fine-grained manner and customizes masked interaction strategy for support-support and query-support relations.
     \item \textbf{To enable longitudinal knowledge accumulation}, we design Inter-task Knowledge Transfer module that retrieves and aggregates relevant temporal knowledge from the temporal knowledge bank, which stores diverse temporal patterns from historical tasks.    
\end{itemize}
\section{Related Work}
\subsection{Few-shot Image Classification}
The goal of few-shot image classification~\citep{fei2006one,pan2024semantic,wang2023few12,ye2021learning,yang2025hyperbolic,yu2023balancing} is to recognize unseen categories with limited annotated samples. Existing few-shot solutions can be broadly categorized into three groups: \textbf{i)} \textit{Augmentation-based} methods typically generate auxiliary data samples by feature augmentation~\citep{chen2019image,chen2019multi,albalak2024improving} or generative models~\citep{li2020adversarial,zhang2018metagan}, to increase the diversity of the support set; \textbf{ii)}  \textit{Optimization-based} methods~\citep{jamal2019task,sun2024meta,rusu2018meta,rajeswaran2019meta,finn2017model} learn good initial model parameters, such that updating the initial parameters via a few gradient steps could adapt well to new tasks; \textbf{iii)} \textit{Metric-based} methods~\citep{yoon2019tapnet,yangone,zhu2023transductive,cheng2023frequency} (\emph{e.g.}, ProtoNet~\citep{snell2017prototypical} and RelationNet~\citep{sung2018learning}) first learn a common feature space for all classes, and then employ pre-defined or learned distance metrics to compare query and support samples. 

Our work shares a similar spirit of metric-based methods~\citep{snell2017prototypical,vinyals2016matching}, which classify query samples by measuring their similarities to support examples in a learned embedding space. However, we address more complex and challenging few-shot action recognition that requires long-range relation modeling between spatio-temporal locations. With respect to this, we further jointly learn inter-video and inter-task relationships to learn distinct temporal features. By unifying such three types of relation modeling under our framework, we can learn task-specific temporal cues for each video from a holistic view.

\subsection{Few-shot Action Recognition}
 Few-shot Action Recognition (FSAR) is gaining much attention due to its practical value in reducing manual annotations at a large scale required by traditional action recognition. 
Compared with few-shot image classification, FSAR is more challenging due to the rich spatial-temporal information in videos. 
Towards FSAR, most existing solutions~\citep{wang2024cross,wu2025efficient,liu2022task,tang2023m3net} belong to the meta-learning paradigm~\citep{snell2017prototypical,vinyals2016matching}, wherein the model learns discriminative temporal features from base action categories, and generalizes learned temporal representations on novel action categories. 
Based on the meta-learning policy, a main group of efforts focus on \textit{temporal representation learning}. 
Early works~\citep{zhu2018compound,zhu2020label} directly aggregate frame-level features to learn a global representation for each video. Though straightforward, these methods neglect complex temporal cues within one video, leading to suboptimal performance.
To address this limitation, recent methods attempt to model temporal relations by, \emph{e.g.}, learning long-range temporal dependencies~\citep{thatipelli2022spatio,wang2022hybrid}, detailed cross-frame patch-level interactions~\citep{zhang2023importance}, exploiting low-level spatial features via feature architecture search~\citep{xing2023revisiting}, or multi-modal feature fusion~\citep{wanyan2023active,fu2020depth,wang2021semantic,huang2024soap}. 
Apart from investigating video feature representation, another line of work turns to designing various \textit{feature matching strategies}, \emph{e.g.}, frame-level feature alignment~\citep{wang2022hybrid}, tuple-level feature alignment~\citep{perrett2021temporal}, and even multi-level feature alignment~\citep{zheng2022few,wu2025efficient,huang2024matching}. 
For example, OTAM~\citep{cao2020few} performs frame-level feature alignment between videos with the DTW algorithm, which takes the temporal ordering characteristics of videos into account. HyRSM~\citep{wang2022hybrid} designs a novel bidirectional Mean Hausdorff metric to obtain temporal matching scores between support and query videos from the set matching perspective.

Despite promising results have been achieved, they typically learn temporal representations independently for each video, ignoring the relation modeling between video samples and tasks. As a result, they fail to gain a holistic semantic pattern understanding from multi-level perspectives. Though a few works (\emph{e.g.}, GgHM~\citep{boosting} and HyRSM~\citep{wang2022hybrid}) attempt to learn task-specific embeddings by considering inter-task relationships, they only focus on the correlations between samples at the video level, overlooking fine-grained temporal interactions among videos.
In contrast, our HR$^{2}$G-shot \textbf{i)} learns inter-video relationships at fine-grained frame level, and \textbf{ii)} gathers temporal knowledge from historical tasks to achieve inter-task knowledge transfer, thereby capturing task-specific temporal features for each video and ensuring a sufficient understanding of holistic semantic patterns for each task.

{\xl{The previous HM$^2$ framework~\cite{gao2024hierarchical} is related to our HR$^2$G-shot, but defines hierarchy along a different dimension.
HM$^2$ constructs part-, frame-, and segment-level appearance features, captures fast- and slow-term motion dynamics through SFAM, models cross-level feature relations using H-RGCN, and performs multimodal matching with DSCM. Its hierarchy therefore concerns feature granularity and appearance--motion matching within the current episode. 
In contrast, our HR$^2$G-shot defines hierarchy over inter-frame, inter-video, and inter-task relation scopes. ISC performs frame-level cross-video contextualization before matching, while IKT retrieves transferable temporal knowledge from historical training tasks. Thus, the two methods investigate complementary aspects of FSAR.}}
\section{Method}
\subsection{Problem Formulation}
The objective of few-shot action recognition (FSAR) is to classify novel action categories with limited samples per class.
Under the few-shot setting, the model generalizes learned knowledge on the training set $\mathcal{D}_{base}$ to novel testing set $\mathcal{D}_{novel}$, where base classes $\mathcal{C}_{base}$ and novel classes $\mathcal{C}_{novel}$ are disjoint.
Consistent with previous works~\citep{cao2020few,wang2022hybrid,tang2024semantic}, we formulate FSAR problem with episodic training and testing. 
Specifically, each $N$-way $K$-shot episode task involves sampled $K$ labeled videos from each of $N$ different action classes as the support set, and a portion of the remaining samples from $N$ classes as the query set. 
During inference, we randomly sample multiple episodic tasks from $\mathcal{D}_{novel}$ and present the average results over these tasks, in order to comprehensively evaluate the performance of the few-shot model.

\begin{figure*}[!t]
    \centering
    \includegraphics[width=0.99\textwidth]{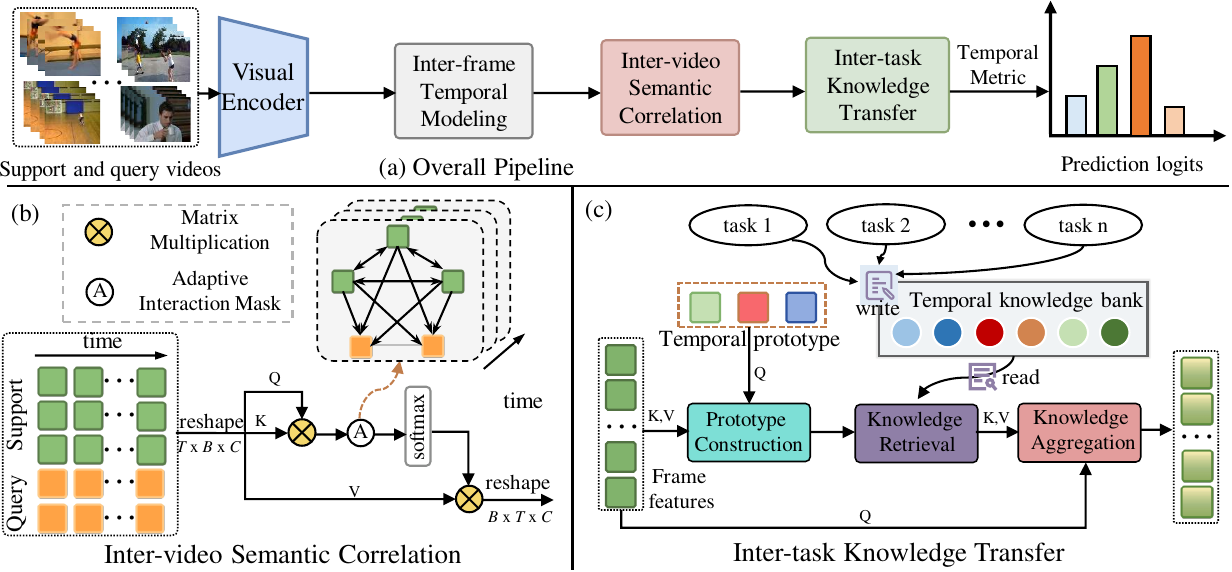}
    \put(-244, 172){\footnotesize(\S\ref{sec::ISC})}
     \put(-161.8, 172){\footnotesize(\S\ref{sec::IKT})}
    \put(-375.6, 204.1){\tiny$\bm{X}_{\mathrm{s}},\bm{X}_{\mathrm{q}}$}
    \put(-286.5, 204.1){\tiny$\bm{F}_{\mathrm{s}},\bm{F}_{\mathrm{q}}$}
    \put(-202.2, 204.1){\tiny$\Tilde{\bm{F}}_{\mathrm{s}},\Tilde{\bm{F}}_{\mathrm{q}}$}
    \put(-109, 190){\footnotesize(Eq.~\ref{eq::temporal_metric})}
     \vspace{-6pt}
    \caption{The overview of HR$^{2}$G-shot. \textbf{(a)} HR$^{2}$G-shot unifies three types of relation modeling (\emph{i.e.}, inter-frame, inter-video, and inter-task) to learn discriminative temporal features. \textbf{(b)} Inter-video Semantic Correlation (ISC) conducts fine-grained cross-video interactions to learn inter-video relationships. \textbf{(c)} To explore inter-task relationships, we retrieve and aggregate temporal knowledge from the bank, which maintains diverse temporal patterns from historical tasks.}
    \label{overview}
    \vspace{-9pt}
\end{figure*}
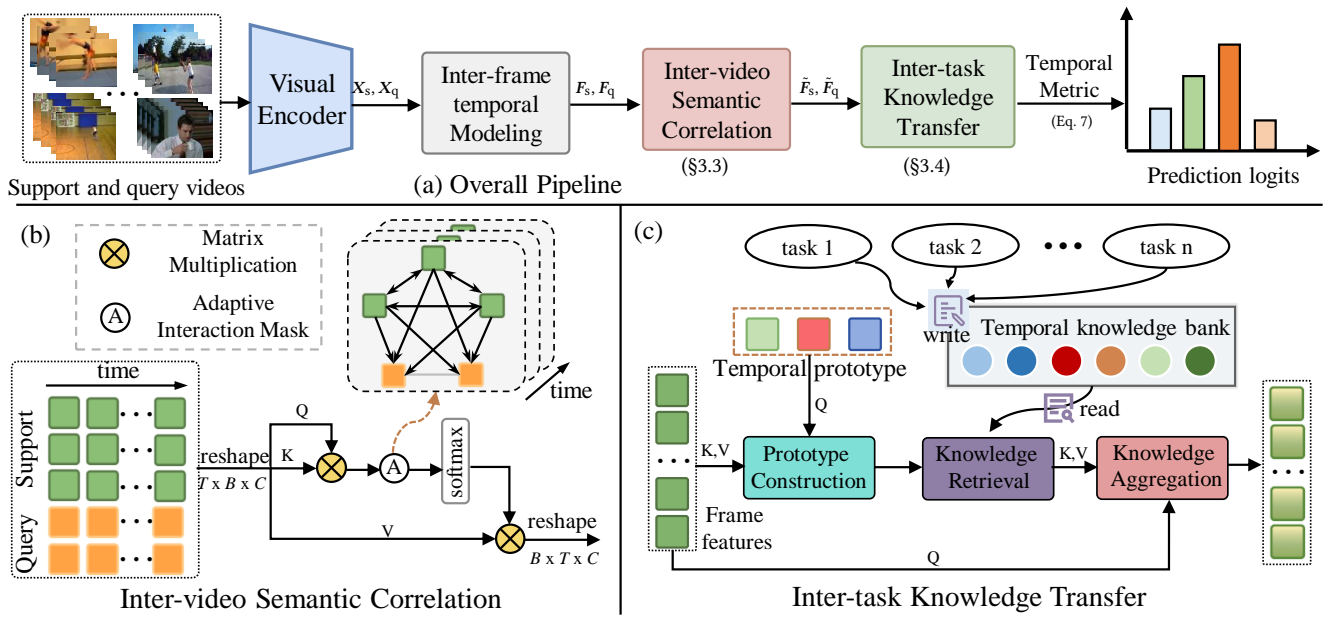
\subsection{Algorithm Overview}
We introduce HR$^{2}$G-shot, which hierarchically models multi-level relations (\emph{i.e.}, inter-frame, inter-video, and inter-task) to capture discriminative andtask-specific temporal patterns from a holistic view for FSAR. The overall architecture of HR$^{2}$G-shot is shown in Fig.~\ref{overview}. Given an $N$-way $K$-shot episode task with $NK$ support videos and $L$ query videos, we employ CLIP visual encoder~\citep{radford2021learning} to get frame-level features for support and query videos, \emph{i.e.}, $\bm{X}_{\mathrm{s}} \in{\mathbb{R}}^{NK\times T \times C} $ and $\bm{X}_{\mathrm{q}} \in {\mathbb{R}}^{L \times T \times C}$, where $T$ is the length of sampled frames in each video. Note that we adopt the Parameter-Efficient Fine-Tuning (PEFT) strategy to fine-tune the visual encoder with minimal trainable parameters as in~\citep{xing2023multimodal,qu2025learning,yangaim}. Next, following previous works~\citep{wang2024clip}, we concatenate frame-level features and corresponding textual features along the temporal dimension and then feed them into the temporal Transformer~\citep{wang2024clip,wang2023molo} (\emph{i.e.}, Inter-frame Temporal Modeling). In this way, we perform inter-frame interactions and obtain temporally enhanced support and query features. To compensate for frame-level relation modeling from other semantic perspectives, we further design two feasible modules: \textbf{i)} Inter-video Semantic Correlation (ISC)~(\S\ref{sec::ISC}) learns inter-video temporal relationships within each task by fine-grained cross-video interactions; \textbf{ii)} Inter-task Knowledge Transfer (IKT) (\S\ref{sec::IKT}) aggregates useful temporal knowledge from historical tasks to learn inter-task relationships. Finally, the obtained relation-enhanced task-specific support and query features, \emph{i.e.}, $\bar{\bm{F}}_{\mathrm{s}}$ and $\bar{\bm{F}}_{\mathrm{q}}$, are fed into temporal matching metrics to get the robust class predictions.

\subsection{Inter-video Semantic Correlation}
\label{sec::ISC}
\begin{figure}[t]
\centering
\includegraphics[width=\columnwidth]{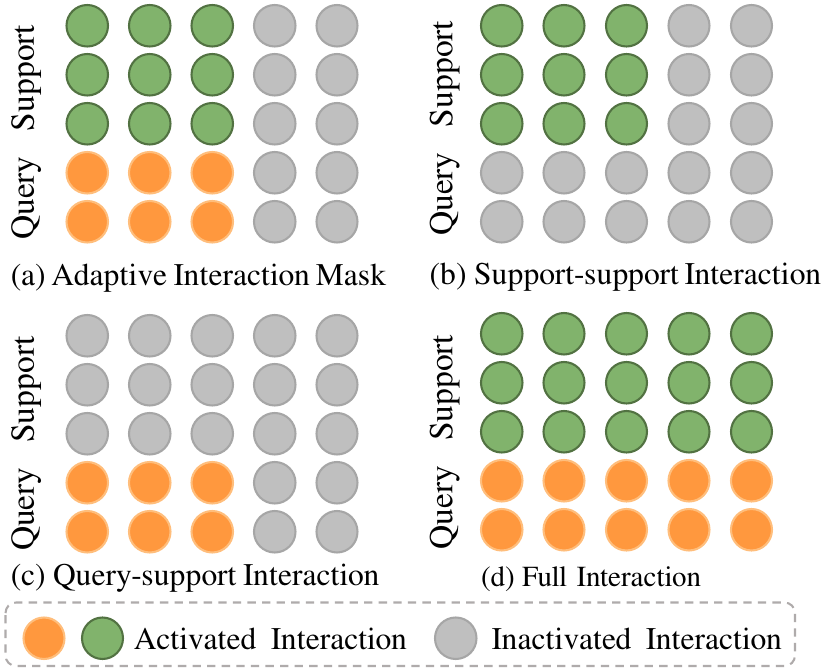}
 \vspace{-10pt}
\caption{Different masked interaction strategies for Inter-video Semantic Correlation.}
\vspace{-12pt}
\label{fig::mask}
\end{figure}

Different from exploiting inter-frame interactions in each video, capturing inter-video relationships can strengthen task-specific semantic patterns for each video. To achieve this goal, we design Inter-video Semantic Correlation (ISC) module to conduct \textbf{Fine-grained Inter-video Interaction} and further learn intra- and inter-class correlation by \textbf{Adaptive Interaction Mask} (Fig.~\ref{fig::mask}(a)) for support-query and support-support videos (Fig.~\ref{overview}(b)). Via Adaptive Interaction Mask, support and query videos aggregate features from different types of videos (\emph{i.e.}, support or query) based on their respective characteristics. Compared with previous works~\citep{wang2022hybrid,boosting}, which only consider utilizing global video features to perform interactions among support and query videos, we attempt to make use of temporal cues in the video to strengthen sufficient temporal pattern understanding among videos.

\noindent\textbf{Fine-grained Inter-video Interaction.} Based on the above analysis, how to perform fine-grained inter-video interactions is crucial for FSAR. A straightforward approach is to perform dense attention along temporal and sample dimensions simultaneously. Thus, given a task with $NK$ support videos and $L$ query videos, the computational complexity of such dense sample attention is $O((NK+L)^{2}T^{2})$, causing high computational costs. Besides, due to the difference between the temporal dimension and the sample dimension, it is not reasonable to treat them equivalently.

Thus, we decouple dense sample attention into inter-frame interaction and our designed Inter-video Semantic Correlation (ISC). The inter-frame interaction is achieved by previous Inter-frame Temporal Modeling, obtaining support and query temporal features, \emph{i.e.}, $\bm{F}_{\mathrm{s}} \in{\mathbb{R}}^{NK\times T \times C} $ and $\bm{F}_{\mathrm{q}} \in {\mathbb{R}}^{L \times T \times C}$. Next, ISC concatenates support and query features, and reshapes these features in one task to ${\mathbb{R}}^{T\times (NK+L) \times C}$. Afterwards, multi-head self-attention along with sample attention matrix $\bm{A} \in{\mathbb{R}}^{(NK+L)\times (NK+L)}$ is adopted to conduct cross-video interactions by attending all frame features at the same temporal location, resulting in task-specific temporal features for support and query videos:
\vspace{-2pt}
\begin{equation}
\label{eq::fine-grained}
         \Tilde{\bm{F}}=[\Tilde{\bm{F}}_{\mathrm{s}},\Tilde{\bm{F}}_{\mathrm{q}}] = \texttt{SelfAttn}(\texttt{Re}(\texttt{Cat}(\bm{F}_{\mathrm{s}},\bm{F}_{\mathrm{q}}))),
\end{equation}
where $\texttt{Re}$ is the reshape operation, and $\texttt{Cat}$ denotes the concatenation operation. $\texttt{SelfAttn}$ represents the standard multi-head self-attention. As such, ISC not only captures task-specific temporal features via interactively transferring temporal knowledge between videos, but also reduces the computational complexity to $O((NK+L)^{2}T)$ (see Table~\ref{tab::fine-attention}).

\noindent\textbf{Adaptive Interaction Mask.} Due to the difference between support and query videos, we propose Adaptive Interaction Mask $\bm{J}\in{\mathbb{R}}^{(NK+L)\times (NK+L)}$ for support-support and query-support interactions, as shown in Fig.~\ref{fig::mask}(a). Specifically in each task, support and query videos only aggregate semantic information from support videos while discarding interactions with query videos by element-wise multiplication of $\bm{A}$ and $\bm{J}$. By transferring knowledge between support-support and query-support videos, support features could learn intra- and inter-class correlation, and query features capture task-specific temporal cues. Besides, as shown in Fig.~\ref{fig::mask}(b)(c)(d), we try other masked interaction strategies for inter-video interactions, including \textit{support-support interaction}, \textit{query-support interaction}, and \textit{full interaction}. Specifically, \textit{support-support interaction} denotes that ISC only considers the interactions among support videos, while \textit{query-support interaction} denotes query videos only aggregate semantic information from support videos in ISC. \textit{Full interaction} means ISC simultaneously considers the interactions among query and support videos, but ignores the distinct characteristics of the query and support videos. The experiments about different strategies are provided in Table~\ref{tab::mask}.

\subsection{Inter-task Knowledge Transfer}
\label{sec::IKT}
Previous FSAR methods typically learn video representations within a single task, failing to explicitly learn transferable temporal knowledge between tasks. Thus, these methods lack generalizability and transferability on unseen action categories. To compensate for this limitation, we design Inter-task Knowledge Transfer (IKT) module to aggregate useful temporal knowledge from the bank, which stores diverse temporal cues shared by different tasks. IKT consists of prototype construction, knowledge retrieval, knowledge aggregation and knowledge bank updating, as shown in Fig.~\ref{overview}(c).

\noindent\textbf{Prototype Construction.} The core idea is to incorporate frame-level features into compact temporal prototypes, so as to filter out redundant information and obtain key temporal cues. We first construct $M$ learnable prototypes $\bm{P} \in {\mathbb{R}}^{M \times C}$ to summarize video representations, where we empirically set $M$ to 3 (ablation study in Table~\ref{tab::prototype}). Given a task consisting of support and query videos, these prototypes adaptively aggregate action dynamic information from support features $\Tilde{\bm{F}}_{\mathrm{s}} \in{\mathbb{R}}^{NK\times T \times C}$:
\vspace{-2pt}
\begin{equation}
         \hat{\bm{P}} = \mathrm{Softmax}(\bm{P}\bm{K}_{\mathrm{t}}^{\top})\bm{V}_{\mathrm{t}} + \bm{P},
\end{equation}
where $\bm{K}_{\mathrm{t}}$ and $\bm{V}_{\mathrm{t}}$ are the linear transformation features of frame-level features $\Tilde{\bm{F}}_{\mathrm{s}}$. As such, such prototypes  $\hat{\bm{P}}$ can adaptively learn useful temporal patterns.

\noindent\textbf{Knowledge Retrieval.} Inspired by~\citep{wu2018unsupervised,zhou2022regional}, we set up a temporal knowledge bank $\bm{M}=[\bm{m}_{1},\bm{m}_{2},...,\bm{m}_{G}] \in {\mathbb{R}}^{G \times C}$ to store task-shared temporal patterns, where $G$ is the number of memory representations and $G$ is empirically set to $50$ (ablation study in Fig.~\ref{fig::banksize}). After obtaining original temporal prototypes $\hat{\bm{P}}$, we perform \textit{knowledge retrieval} via only aggregating semantically related memory from the bank:
\vspace{-2pt}
\begin{equation}
\label{eq::topk}
         \bm{P}^\prime= \sum_{g=1}^G \mathrm{topk}(\hat{\bm{P}},\bm{m}_{g})\bm{m}_{g},
\end{equation}
where $\mathrm{topk}$ is the function that retrieves $O$ most similar prototypes from the bank according to the semantic correspondences. We take a hyper-parameter $\kappa \!=\! O/G$ to determine the number of retrieved memory representations. $\bm{P}^\prime$ is retrieved historical knowledge related to the current task, which could be used to further enhance task-specific video representations.

\noindent\textbf{Knowledge Aggregation.} Knowledge Aggregation aims to make use of learned historical knowledge to enhance temporal features. Specifically, we first deliver the union of original temporal prototypes $\hat{\bm{P}}$ and retrieved memory $\bm{P}^\prime$ by element-wise addition to obtain task-specific prototypes $\bar{\bm{P}}$. Then, support features $\Tilde{\bm{F}}_{\mathrm{s}}$ aggregate temporal knowledge provided by prototypes $\bar{\bm{P}}$ via cross-attention mechanism:
\begin{equation}\small
         \bar{\bm{F}}_{\mathrm{s}} = \texttt{CrossAttn}(\Tilde{\bm{F}}_{\mathrm{s}},\bar{\bm{P}}),
\end{equation}
where $\texttt{CrossAttn}$ is the standard cross-attention operation, and $\bar{\bm{F}}_{\mathrm{s}}$ is enhanced temporal features of $\Tilde{\bm{F}}_{\mathrm{s}}$ by exploring inter-task relationships. By aggregating temporal knowledge from previous tasks, we strengthen discriminative dynamic patterns for each new episodic task.

\noindent\textbf{Knowledge Bank Updating.} At each episodic task, the temporal knowledge bank is continually updated to involve new learned knowledge from the current task. Our updating strategy is as follows: given $i$-th temporal prototype $\bm{p}_{i}$ in $\hat{\bm{P}}$, if the knowledge bank is not full, the prototype is directly assigned; if the bank is full, we first compute the cosine similarity between current prototype and $G$ memory representations in $\bm{M}$, and update the most similar one as follows: 
\begin{equation}
     \bm{m}_{t} = \texttt{SelectMax}\left( \frac{\bm{p}_{i}\cdot\bm{m}_{j}}{||\bm{p}_{i}||||\bm{m}_{j}||} \right), \bm{m}_{j} \in \bm{M},
\end{equation}
\begin{equation}
\label{eq:miu}
     \bm{m}_{t}\gets \mu\bm{m}_{t} + (1-\mu)\bm{p}_{i},
\end{equation}
where $\texttt{SelectMax}$ denotes the function that selects most similar memory representation from $\bm{M}$ as $\bm{m}_{t}$, and $\mu\!\in\![0,1]$  is the momentum for memory evolution. Note that during the testing phase, the temporal knowledge bank remains unchanged and does not store any information from testing data, avoiding information leakage.

\noindent\textbf{\textit{More In-Depth Discussion}.} We emphasize that our designed Inter-task Knowledge Transfer (IKT) module strictly adheres to the standard few-shot learning protocol, and no information leakage occurs during testing. Specifically, the temporal knowledge bank in IKT only stores learned knowledge from support samples of training tasks and remains fixed during the testing phase. When performing knowledge retrieval at test time, our approach only accesses the support and query videos of the current task, and the knowledge bank used at test time remains unchanged and does \textit{not} incorporate any features from test-time videos (\emph{i.e.}, unseen classes). 

In addition, our method learns useful temporal knowledge by capturing generalizable and task-shared semantic patterns during training, rather than introducing class-specific feature leakage from testing data. This is aligned with the goal of few-shot learning, which is to generalize to new classes by utilizing prior knowledge.

%
\subsection{Training and Testing with HR$^{2}$G-shot}
\noindent\textbf{Training Objective.} After jointly exploring three types of relation modeling, we obtain support features $\bar{\bm{F}}_{\mathrm{s}}$ and query features $\bar{\bm{F}}_{\mathrm{q}}$. Given $i$-th  support frame-level features $\bar{\bm{f}}_{\mathrm{s}}^{i} \in{\mathbb{R}}^{ T \times C}$ in $\bar{\bm{F}}_{\mathrm{s}}$  and $j$-th query frame-level features $\bar{\bm{f}}_{\mathrm{q}}^{j} \in{\mathbb{R}}^{T \times C}$ in $\bar{\bm{F}}_{\mathrm{q}}$, like in previous methods~\citep{cao2020few,wang2024clip}, we adopt the temporal alignment metric to obtain query-support distances: 
\begin{equation}
\label{eq::temporal_metric}
{\mathcal{D}} = \mathtt{Metric}(\bar{\bm{f}}_{\mathrm{q}}^{j}, \bar{\bm{f}}_{\mathrm{s}}^{i}),
\end{equation}
where $\mathtt{Metric}$ denotes the  OTAM~\citep{cao2020few} metric by default. Then we can use the output support-query distances as logits to compute cross-entropy loss $\mathcal{L}_{\mathrm{CE}}$ over the ground-truth labels. 

\noindent\textbf{Testing.} During testing, we freeze the entire framework, and directly extract support and query features in a single feedforward pass for the unseen test classes. Note that the temporal knowledge bank is also frozen, not incorporating any features from test-time videos. 
Finally, we can utilize the obtained distance in Eq.~\ref{eq::temporal_metric} as logits to produce query predictions. 
\begin{table*}[!t]
\caption{\textbf{Quantitative comparison results on HMDB51~\cite{kuehne2011hmdb} and UCF101~\cite{soomro2012ucf101}} (see \S\ref{sec::sota}). The experiments are conducted under the $5$-way $K$-shot setting. ``INet-RN50'' denotes ResNet-50 pre-trained on ImageNet. We highlight \textbf{best} and \underline{second-best} results.}
\vspace{-5pt}
\label{tab:tab1}
\centering
\small
\resizebox{0.99\textwidth}{!}{
\setlength\tabcolsep{8pt}
\renewcommand\arraystretch{1.2}
\begin{tabular}{rl||c|ccc|ccc}
\thickhline
\rowcolor{mygray} &&
& \multicolumn{3}{c|}{HMDB51} 
& \multicolumn{3}{c}{UCF101} \\ 
\rowcolor{mygray}
\multicolumn{2}{c||}{\multirow{-2}{*}{Method}} 
& \multicolumn{1}{c|}{\multirow{-2}{*}{\hspace{0.5mm} Pre-training \hspace{0.5mm}}}  
& $1$-shot (\%) & $3$-shot (\%) & $5$-shot (\%) 
& $1$-shot (\%) & $3$-shot (\%) & $5$-shot (\%) \\
\hline
\hline
AmeFuNet~\cite{fu2020depth}\!\!\!\!\!\! 
& \!\!\pub{MM20} 
& INet-RN50 
& $60.2$ & $-$ & $75.5$ 
& $85.1$ & $-$ & $95.5$ \\

TRX~\cite{perrett2021temporal}\!\!\!\!\!\! 
& \!\!\pub{CVPR21} 
& INet-RN50 
& $53.1$ & $66.8$ & $75.6$ 
& $78.2$ & $92.4$ & $96.1$ \\ 

TA$^{2}$N~\cite{ta2n}\!\!\!\!\!\! 
& \!\!\pub{AAAI22} 
& INet-RN50 
& $59.7$ & $-$ & $73.9$ 
& $81.9$ & $-$ & $95.1$ \\

MTFAN~\cite{wu2022motion}\!\!\!\!\!\! 
& \!\!\pub{CVPR22} 
& INet-RN50 
& $59.0$ & $-$ & $74.6$ 
& $84.8$ & $-$ & $95.1$ \\

HyRSM~\cite{wang2022hybrid}\!\!\!\!\!\! 
& \!\!\pub{CVPR22} 
& INet-RN50 
& $60.3$ & $71.7$ & $76.0$ 
& $83.9$ & $93.0$ & $94.7$ \\

STRM~\cite{thatipelli2022spatio}\!\!\!\!\!\! 
& \!\!\pub{CVPR22} 
& INet-RN50 
& $52.3$ & $67.4$ & $77.3$ 
& $80.5$ & $92.7$ & $96.9$ \\

HCL~\cite{zheng2022few}\!\!\!\!\!\! 
& \!\!\pub{ECCV22} 
& INet-RN50 
& $59.1$ & $71.2$ & $76.3$ 
& $82.6$ & $91.0$ & $94.5$ \\

SloshNet~\cite{xing2023revisiting}\!\!\!\!\!\! 
& \!\!\pub{AAAI23} 
& INet-RN50 
& $-$ & $-$ & $77.5$ 
& $-$ & $-$ & $97.1$ \\

MoLo~\cite{wang2023molo}\!\!\!\!\!\! 
& \!\!\pub{CVPR23} 
& INet-RN50 
& $60.8$ & $72.0$ & $77.4$ 
& $86.0$ & $93.5$ & $95.5$ \\

GgHM~\cite{boosting}\!\!\!\!\!\! 
& \!\!\pub{ICCV23} 
& INet-RN50 
& $61.2$ & $-$ & $76.9$ 
& $85.2$ & $-$ & $96.3$ \\

{\xl {HM$^2$}}~\cite{gao2024hierarchical}\!\!\!\!\!\! 
& \!\!\pub{TMM24} 
& {\xl{ INet-RN50}}
& {\xl {$61.8$}} & {\xl {$72.6$}} & {\xl {$77.9$}} 
& {\xl{$86.6$}} & {\xl{ $94.4$}} & {\xl{ $96.2$}} \\

CLIP-FSAR~\cite{wang2024clip}\!\!\!\!\!\! 
& \!\!\pub{IJCV24} 
& CLIP-RN50 
& $69.2$ & $77.6$ & $80.3$ 
& $91.3$ & $95.1$ & $97.0$ \\

A$^{2}$M$^{2}$-Net~\cite{gao20252}\!\!\!\!\!\! 
& \!\!\pub{IJCV25} 
& INet-RN50 
& $61.8$ & $73.4$ & $76.6$ 
& $86.1$ & $94.0$ & $95.8$ \\

CLIP-Freeze~\cite{radford2021learning}\!\!\!\!\!\! 
& \!\!\pub{ICML21} 
& CLIP-ViT-B 
& $58.2$ & $72.7$ & $77.0$ 
& $89.7$ & $94.3$ & $95.7$ \\

CapFSAR~\cite{wang2023few}\!\!\!\!\!\! 
& \!\!\pub{ArXiv23} 
& CLIP-ViT-B 
& $65.2$ & $-$ & $78.6$ 
& $93.3$ & $-$ & $97.8$ \\

MVP-shot~\cite{qu2024mvp}\!\!\!\!\!\! 
& \!\!\pub{TMM25} 
& CLIP-ViT-B 
& $77.0$ & $-$ & \underline{$88.1$} 
& $96.8$ & $-$ & $99.0$ \\

CLIP-FSAR~\cite{wang2024clip}\!\!\!\!\!\! 
& \!\!\pub{IJCV24} 
& CLIP-ViT-B 
& $77.1$ & \underline{$84.1$} & $87.7$ 
& $97.0$ & \underline{$98.4$} & $99.1$ \\

EMP-Net~\cite{wu2025efficient}\!\!\!\!\!\! 
& \!\!\pub{ECCV24} 
& CLIP-ViT-B 
& $76.8$ & $-$ & $85.8$ 
& $94.3$ & $-$ & $98.2$ \\

TEAM~\cite{lee2025temporal}\!\!\!\!\!\! 
& \!\!\pub{CVPR25} 
& INet-ViT-B 
& $70.9$ & $-$ & $85.5$ 
& $94.5$ & $-$ & $98.8$ \\
MA-FSAR~\cite{xing2023multimodal}\!\!\!\!\!\! 
& \!\!\pub{PR26} 
& CLIP-ViT-B 
& \underline{$83.4$} & $-$ & $87.9$ 
& \underline{$97.2$} & $-$ & \underline{$99.2$} \\

\hline
\hline

\multicolumn{2}{c||}{\textbf{HR$^{2}$G-shot (Ours)}} 
& CLIP-ViT-B 
& $\mathbf{85.6}$ & $\mathbf{87.9}$ & $\mathbf{88.6}$ 
& $\mathbf{98.0}$ & $\mathbf{99.0}$ & $\mathbf{99.3}$ \\

\hline
\end{tabular}
}
\vspace{-6pt}
\end{table*}
\begin{table*}[!t]
\caption{\textbf{Quantitative comparison results on Kinetics~\cite{carreira2017quo}, SSv2-full~\cite{goyal2017something}, and SSv2-small~\cite{goyal2017something}} (see \S\ref{sec::sota}). The experiments are conducted under the $5$-way $K$-shot setting. ``INet-RN50'' denotes ResNet-50 pre-trained on ImageNet. We highlight \textbf{best} and \underline{second-best} results.}
\vspace{-5pt}
\label{tab:tab2}
\centering
\small
\resizebox{0.98\textwidth}{!}{
\setlength\tabcolsep{8pt}
\renewcommand\arraystretch{1.2}
\begin{tabular}{rl||c|cc|cc|cc}
\thickhline
\rowcolor{mygray}
&& & \multicolumn{2}{c|}{Kinetics} 
& \multicolumn{2}{c|}{SSv2-full}
& \multicolumn{2}{c}{SSv2-small} \\
\rowcolor{mygray}
\multicolumn{2}{c||}{\multirow{-2}{*}{Method}} 
& \multicolumn{1}{c|}{\multirow{-2}{*}{Pre-training}} 
& $1$-shot (\%) & $5$-shot (\%) 
& $1$-shot (\%) & $5$-shot (\%) 
& $1$-shot (\%) & $5$-shot (\%) \\
\hline 
\hline

CMN~\cite{zhu2018compound}\!\!\!\!\!\!
& \!\!\pub{ECCV18} 
& INet-RN50 
& $60.5$ & $78.9$ 
& $36.2$ & $48.9$ 
& $34.4$ & $43.8$ \\

OTAM~\cite{cao2020few}\!\!\!\!\!\!
& \!\!\pub{CVPR20} 
& INet-RN50 
& $73.0$ & $85.8$ 
& $42.8$ & $52.3$ 
& $-$ & $-$ \\

AmeFuNet~\cite{fu2020depth}\!\!\!\!\!\!
& \!\!\pub{MM20} 
& INet-RN50 
& $74.1$ & $86.8$ 
& $-$ & $-$ 
& $-$ & $-$ \\

TRX~\cite{perrett2021temporal}\!\!\!\!\!\!
& \!\!\pub{CVPR21} 
& INet-RN50 
& $63.6$ & $85.9$ 
& $-$ & $64.6$ 
& $-$ & $59.1$ \\

TA$^{2}$N~\cite{ta2n}\!\!\!\!\!\!
& \!\!\pub{AAAI22} 
& INet-RN50 
& $72.8$ & $85.8$ 
& $47.6$ & $61.0$ 
& $-$ & $-$ \\

MTFAN~\cite{wu2022motion}\!\!\!\!\!\!
& \!\!\pub{CVPR22} 
& INet-RN50 
& $74.6$ & $87.4$ 
& $45.7$ & $60.4$ 
& $-$ & $-$ \\

HyRSM~\cite{wang2022hybrid}\!\!\!\!\!\!
& \!\!\pub{CVPR22} 
& INet-RN50 
& $73.7$ & $86.1$ 
& $54.3$ & $69.0$ 
& $40.6$ & $56.1$ \\

STRM~\cite{thatipelli2022spatio}\!\!\!\!\!\!
& \!\!\pub{CVPR22} 
& INet-RN50 
& $62.9$ & $86.7$ 
& $43.1$ & $68.1$ 
& $37.1$ & $55.3$ \\

HCL~\cite{zheng2022few}\!\!\!\!\!\!
& \!\!\pub{ECCV22} 
& INet-RN50 
& $73.7$ & $85.8$ 
& $47.3$ & $64.9$ 
& $38.9$ & $55.4$ \\

SloshNet~\cite{xing2023revisiting}\!\!\!\!\!\!
& \!\!\pub{AAAI23} 
& INet-RN50 
& $-$ & $87.0$ 
& $-$ & $-$ 
& $-$ & $-$ \\

MoLo~\cite{wang2023molo}\!\!\!\!\!\!
& \!\!\pub{CVPR23} 
& INet-RN50 
& $74.0$ & $85.6$ 
& $55.0$ & $69.6$ 
& $41.9$ & $56.2$ \\

GgHM~\cite{boosting}\!\!\!\!\!\!
& \!\!\pub{ICCV23} 
& INet-RN50 
& $74.9$ & $87.4$ 
& $54.5$ & $69.2$ 
& $-$ & $-$ \\

{\xl {HM$^2$}}~\cite{gao2024hierarchical}\!\!\!\!\!\!
& \!\!\pub{TMM24} 
& {\xl{ INet-RN50}} 
& {\xl{ $75.5$}} & {\xl{ $87.5$}} 
& {\xl{ $-$}} & {\xl{ $-$}} 
& {\xl{ $42.5$}} & {\xl{ $56.5$}} \\

CLIP-FSAR~\cite{wang2024clip}\!\!\!\!\!\!
& \!\!\pub{IJCV24} 
& CLIP-RN50 
& $87.6$ & $91.9$ 
& $58.1$ & $62.8$ 
& $52.0$ & $55.8$ \\

A$^{2}$M$^{2}$-Net~\cite{gao20252}\!\!\!\!\!\!
& \!\!\pub{IJCV25} 
& INet-RN50 
& $74.5$ & $87.5$ 
& $56.7$ & \underline{$74.1$} 
& $42.9$ & $60.5$ \\

CLIP-Freeze~\cite{radford2021learning}\!\!\!\!\!\!
& \!\!\pub{ICML21} 
& CLIP-ViT-B 
& $78.9$ & $91.9$ 
& $30.0$ & $42.4$ 
& $29.5$ & $42.5$ \\

CapFSAR~\cite{wang2023few}\!\!\!\!\!\!
& \!\!\pub{ArXiv23} 
& CLIP-ViT-B 
& $84.9$ & $93.1$ 
& $51.9$ & $68.2$ 
& $45.9$ & $59.9$ \\

MVP-shot~\cite{qu2024mvp}\!\!\!\!\!\!
& \!\!\pub{TMM25} 
& CLIP-ViT-B 
& $91.0$ & $95.1$ 
& $-$ & $-$ 
& $55.4$ & $62.0$ \\

CLIP-FSAR~\cite{wang2024clip}\!\!\!\!\!\!
& \!\!\pub{IJCV24} 
& CLIP-ViT-B 
& $94.8$ & $95.4$ 
& $62.1$ & $72.1$ 
& $54.6$ & $61.8$ \\

EMP-Net~\cite{wu2025efficient}\!\!\!\!\!\!
& \!\!\pub{ECCV24} 
& CLIP-ViT-B 
& $89.1$ & $93.5$ 
& $63.1$ & $73.0$ 
& $57.1$ & \underline{$65.7$} \\

TEAM~\cite{lee2025temporal}\!\!\!\!\!\!
& \!\!\pub{CVPR25} 
& INet-ViT-B 
& $83.3$ & $92.9$ 
& $-$ & $-$ 
& $47.2$ & $63.1$ \\

MA-FSAR~\cite{xing2023multimodal}\!\!\!\!\!\!
& \!\!\pub{PR26} 
& CLIP-ViT-B 
& $\mathbf{95.7}$ & \underline{$96.0$} 
& \underline{$63.3$} & $72.3$ 
& \underline{$59.1$} & $64.5$ \\

\hline 
\hline

\multicolumn{2}{c||}{\textbf{HR$^{2}$G-shot (Ours)}} 
& CLIP-ViT-B 
& \underline{$95.2$} & $\mathbf{96.4}$ 
& $\mathbf{65.4}$ & $\mathbf{74.8}$ 
& $\mathbf{60.2}$ & $\mathbf{66.0}$ \\

\hline
\end{tabular}}
\vspace{-6pt}
\end{table*}

\section{Experiments}
\subsection{Experimental Setup}
\sssection{Dataset.} We evaluate our HR$^{2}$G-shot on five commonly used datasets, \emph{i.e.}, Kinetics~\citep{carreira2017quo}, SSv2-small~\citep{goyal2017something}, SSv2-full~\citep{goyal2017something}, HMDB51~\citep{kuehne2011hmdb}, and UCF101~\citep{soomro2012ucf101}. 
For HMDB51 and UCF101, we adopt the few-shot split from~\citep{zhang2020few,bishaytarn}, with $31$/$10$/$10$ classes and $70$/$10$/$21$ classes for \texttt{train}/\texttt{val}/\texttt{test}, respectively. 
Following previous works~\citep{wang2023molo,xia2023few,cao2022searching}, we divide Kinetics, SSv2-small and SSv2-full into three splits: $64$/$12$/$24$ classes used for \texttt{train}, \texttt{val}, and \texttt{test}, respectively. The difference between SSv2-small and SSv2-full is that SSv2-full contains more videos per class in \texttt{train} set.

\sssection{Evaluation.} Following  standard evaluation protocols~\citep{cao2020few,boosting,wang2023molo,liu2023lite,yu2023multi}, $5$-way $1$-shot  and  $5$-shot accuracy over $10,000$ tasks are used for evaluation.

\subsection{Implementation Details}
\sssection{Network Architecture.}  HR$^{2}$G-shot adopts pre-trained CLIP ViT-B~\citep{radford2021learning} as our backbone for parameter-efficient fine-tuning, for a fair comparison with previous methods~\citep{qu2024mvp,wang2024clip,xing2023multimodal}. In IKT, we empirically set the size of the temporal knowledge bank $G\! = \!50$ (see Fig.~\ref{fig::banksize}). By default, the number of temporal prototypes $M$ is set to $3$ (see Table~\ref{tab::prototype}). For other hyper-parameters, we  empirically set retrieval ratio $\kappa$ and momentum $\mu$ to $0.7$ (see Table~\ref{tab::retrieval}) and $0.99$ (see Table~\ref{tab::momentum}), respectively.

\sssection{Network Training.} In line with previous works~\citep{cao2020few,huang2022compound,wang2024clip,boosting}, we uniformly and sparsely sample $T \!=\! 8$ frames from each video to encode frame-level representations,  resizing each frame to a height of 256 pixels. 
During the training phase, we apply standard data augmentation techniques, including random horizontal flipping, cropping, and color jittering. Conversely, only center cropping is utilized during the testing phase to ensure deterministic evaluation.
We freeze both the CLIP visual encoder and text encoder, and only finetune lightweight adapters in the visual encoder as~\citep{xing2023multimodal}.
The model is optimized using the Adam~\citep{kingma2014adam} optimizer, coupled with a multi-step learning rate scheduler. The entire training process spans $10$ epochs.

\sssection{Reproducibility.} All experiments are conducted on three NVIDIA 3090 GPUs with 24GB memory in PyTorch. To ensure reproducibility, full code will be released.
\begin{table}
\caption{The impact and parameters of core components on SSv2-small~\citep{goyal2017something} and HMDB51~\citep{kuehne2011hmdb} under the 5-way 1-shot setting~(see \S\ref{ex::abs}).}
\vspace{-5pt}
\label{core_module}
\centering
\small
\resizebox{0.49\textwidth}{!}{
\setlength\tabcolsep{3pt}
\renewcommand\arraystretch{0.95}
\begin{tabular}{c|clcc}
\thickhline
\rowcolor{mygray}
 Method Component &  Model Params&& SSv2-small & HMDB51  \\ \hline  \hline
  \textsc{Baseline}    &   $100.8$M &  & $55.0$ & $81.2$     \\  \arrayrulecolor{gray}\hdashline\arrayrulecolor{black}
       ISC \textit{only}  & $103.0$M&\!\!\!\!\!\!\!\!\!\!\!\!\!\!  \textbf{{(+2.2)}}      & $58.7$ & $84.0$    \\ 
   IKT \textit{only}   & $104.3$M&\!\!\!\!\!\!\!\!\!\!\!\!\!\! \textbf{{(+3.5)}}    & $57.8$   & $83.2$      \\ 
   \arrayrulecolor{gray}\hdashline\arrayrulecolor{black}
 HR$^{2}$G-shot \textbf{(Ours)}   & $106.5$M&\!\!\!\!\!\!\!\!\!\!\!\!\!\! \textbf{{(+5.7)}}       & $\mathbf{60.2}$ & $\mathbf{85.6}$     \\ \hline
\end{tabular}}  \vspace{-12pt}
\end{table}
 
\subsection{Comparison with State-of-the-Arts}
\label{sec::sota}
Table~\ref{tab:tab1} and \ref{tab:tab2} illustrate our compelling results over the top-leading FSAR solutions on five datasets (\emph{i.e.}, SSv2-small~\citep{goyal2017something}, SSv2-full~\citep{goyal2017something}, HMDB51~\citep{kuehne2011hmdb}, UCF101~\citep{soomro2012ucf101}, and Kinetics~\citep{carreira2017quo}). Our method reports the results on CLIP-ViT-B visual encoder. 
For spatial-related datasets, our method yields remarkable performance on most task settings across HMDB51, UCF101, and Kinetics. 
Especially for 1-shot tasks, it surpasses the previous SOTA (\emph{i.e.}, MA-FSAR~\citep{xing2023multimodal}) by  2.2\% on HMDB51 and 0.8\% on UCF101, respectively. 
Temporal-related datasets (SSv2-small and SSv2-full) require models to comprehend complex temporal information, making it much more challenging. 
Our approach still achieves dominant results, surpassing other
competitors across all metric task settings. This reinforces our belief that our method effectively makes use of the temporal knowledge bank, which collects transferable temporal cues from historical tasks, to yield discriminative temporal features.
All of the above improvements across all datasets and task settings show our HR$^{2}$G-shot has strong generalization for different scenes.
We attribute this to the fact that we learn task-specific temporal features by transferring knowledge at both the video-level and task-level.

\subsection{Diagnostic Experiment}
\label{ex::abs}
\subsubsection{Key Component Analysis} 
\label{sec:key}
We first investigate the effectiveness of each component in HR$^{2}$G-shot, \emph{i.e.}, Inter-video Semantic Correlation (ISC) and  Inter-task Knowledge Transfer (IKT), which is summarized in Table~\ref{core_module}. First, our proposed ISC leads to $3.7\%$ and $2.8\%$  performance gains against the baseline on SSv2-small~\citep{goyal2017something} and HMDB51~\citep{kuehne2011hmdb}, respectively, demonstrating the value of inter-video relation modeling. Second, after incorporating our proposed IKT into the baseline, our method improves on the two datasets by $2.8\%$ and $2.0\%$, verifying that capturing task-shared temporal knowledge by exploring inter-task relationships can yield task-specific video representations. Third, our full model HR$^{2}$G-shot achieves the best performance, confirming the joint effectiveness of our overall algorithm design. Different from the reported learnable model parameters in Table~\ref{efficiency}, we provide full model parameters in Table~\ref{core_module}.  As seen, our algorithm brings a modest amount of extra parameters, while leveraging such a performance leap.

\begin{table}[t]
\caption{Ablation study on different inter-video interaction manners in ISC on SSv2-small~\citep{goyal2017something} and HMDB51~\citep{kuehne2011hmdb} under the 5-way 1-shot setting~(see \S\ref{ex::abs}).}
\vspace{-5pt}
\label{tab::fine-attention}
\centering
\small
\resizebox{0.49\textwidth}{!}{
\setlength\tabcolsep{3pt}
\renewcommand\arraystretch{1.0}
\begin{tabular}{c|ccc}
\thickhline
\rowcolor{mygray}
 Method  & FLOPs (M)& SSv2-small & HMDB51  \\ \hline  \hline
  Video-level global attention    & $53$    & $56.5$ & $83.6$     \\  
     Frame-level  dense attention  & $252$      & $58.9$ & $84.2$    \\ 
 \textbf{Ours}      & $58$    & $\mathbf{60.2}$ & $\mathbf{85.6}$     \\ \hline
\end{tabular}} \vspace{-12pt}
\end{table}

\subsubsection{Impact of Inter-video Interaction Manner in ISC} 
\label{sec:inter}
We study the impact of our fine-grained inter-video interaction (Eq.~\ref{eq::fine-grained}) by contrasting it with video-level global attention~\citep{wang2022hybrid} and frame-level dense attention. ``video-level global attention" refers to conducting inter-video interaction via global video features, and ``frame-level dense attention" means one frame token interacts with all other frame tokens in the current task. As outlined in Table~\ref{tab::fine-attention}, our fine-grained inter-video interaction proves to be \textit{effective}--it outperforms video-level global attention and frame-level dense attention by $3.7\%$ and $1.3\%$ on SSv2-small respectively, and \textit{efficient}--its FLOPs are significantly less than frame-level dense attention. The reason why our fine-grained interaction outperforms frame-level dense attention may be that frame-level dense attention often introduces redundant or noisy interactions. Thus, we conclude that our fine-grained inter-video interaction achieves a trade-off between effectiveness and efficiency for capturing task-specific temporal cues.
\newcommand{\UnifiedResizeWidth}{\columnwidth}
\begin{table}[t]
\centering
\caption{Ablation study  on different masked interaction strategies under the $5$-way $1$-shot setting on SSv2-small~\citep{goyal2017something} and HMDB51~\citep{kuehne2011hmdb}~(see \S\ref{ex::abs}).}
\vspace{-5pt}
\label{tab::mask}
\setlength{\tabcolsep}{11pt}   
\renewcommand\arraystretch{1.0}
\resizebox{0.49\textwidth}{!}{\begin{tabular}{c|cc}
\thickhline
\rowcolor{mygray}
Masked Interaction Strategy& SSv2-small & HMDB51 \\ \hline \hline
support-support & $58.7$ & $84.7$ \\
query-support   & $59.0$ & $84.5$ \\
full            & $57.8$ & $84.0$ \\
\textbf{Ours} & $\mathbf{60.2}$ & $\mathbf{85.6}$ \\ \hline
\end{tabular}}\vspace{-3pt}
\end{table}

\begin{table}[t]
\centering
\caption{Impact of different retrieval ratio $\kappa$ in IKT under the $5$-way $1$-shot setting on SSv2-small~\citep{goyal2017something} and HMDB51~\citep{kuehne2011hmdb}~(see \S\ref{ex::abs}). The adopted design is marked in \textcolor{red}{red}.}
\vspace{-5pt}
\label{tab::retrieval}
\setlength{\tabcolsep}{14pt}   
\renewcommand\arraystretch{1.0}
\resizebox{\columnwidth}{!}{
\begin{tabular}{c|cc}
\thickhline
\rowcolor{mygray}
 Retrieval Ratio $\kappa$ & SSv2-small & HMDB51  \\ \hline  \hline
  $\kappa=0.3$         & $58.9$   & $84.3$   \\  
       $\kappa=0.5$        & $59.5$   & $84.9$  \\ 
   \textcolor{red}{$\kappa=0.7$}     & $\mathbf{60.2}$   & $\mathbf{85.6}$     \\ 
$\kappa=0.9$      & $59.9$    & $85.3$     \\ 
 $\kappa=1.0$      & $59.7$    & $85.2$     \\       \hline
\end{tabular}}\vspace{-12pt}
\end{table}

\begin{table}[t]
\centering
\caption{Impact of temporal prototype number $M$ in IKT under the $5$-way $1$-shot setting on SSv2-small~\citep{goyal2017something} and HMDB51~\citep{kuehne2011hmdb}~(see \S\ref{ex::abs}). The adopted design is marked in \textcolor{red}{red}.}
\vspace{-5pt}
\label{tab::prototype}
\resizebox{\columnwidth}{!}{
\setlength{\tabcolsep}{10pt}   
\renewcommand\arraystretch{1.0}
\begin{tabular}{c|cc}
\thickhline
\rowcolor{mygray}
 Prototype Number $M$ & SSv2-small  & HMDB51 \\ \hline  \hline
  $M=1$         & $59.0$    & $84.7$    \\  
       $M=2$        & $59.5$   & $85.0$    \\ 
   \textcolor{red}{$M=3$}      & $\mathbf{60.2}$   & $\mathbf{85.6}$       \\ 
$M=4$      & $59.6$     & $85.3$      \\ 
 $M=8$      & $58.8$      & $85.2$     \\       \hline
\end{tabular}}\vspace{-3pt}
\end{table}

\subsubsection{Impact of Different Masked Interaction Strategies in ISC}
\label{sec:mask}
We further analyze the influence of our masked interaction strategy on SSv2-small~\citep{goyal2017something} and HMDB51~\citep{kuehne2011hmdb}. Fig.~\ref{fig::mask}(b)(c)(d) shows three other alternative masked strategies, \emph{i.e.}, support-support interaction, query-support interaction and full interaction. As reported in Table~\ref{tab::mask}, our masked interaction strategy (\emph{i.e.}, adaptive Interaction Mask) yields significant performance advancements compared with other alternatives.
Notably, on the challenging SSv2-small benchmark, our Adaptive Interaction Mask surpasses the second-best strategy (query-support) by 1.2\% and the full interaction baseline by 2.4\%. It is worth noting that the full interaction strategy yields suboptimal results, indicating that simply aggregating all token interactions leads to information redundancy. Our approach addresses this by dynamically tailoring the interaction mask, thereby facilitating more precise cross-video reasoning.

\subsubsection{Impact of Different Retrieval Ratio $\kappa$ in IKT} 
\label{sec:ratio}
We next investigate the impact of retrieval ratio $\kappa$ (Eq.~\ref{eq::topk}), which controls the number of retrieved memory representations from the knowledge bank. Table~\ref{tab::retrieval} provides comparison results with regard to different retrieval ratios under the 5-way 1-shot setting on SSv2-small~\citep{goyal2017something} and HMDB51~\citep{kuehne2011hmdb}. We can clearly observe that, our algorithm performs best with a relatively large retrieval ratio (\emph{i.e.}, $\kappa=0.7$). This verifies that we discard the semantically unrelated temporal knowledge in the bank to alleviate the negative effects of meaningless semantic information. 
Specifically, the accuracy improves as $\kappa$ increases from $0.3$ to $0.7$, achieving the optimal performance at $\kappa=0.7$ on both benchmarks. This improvement suggests that a broader retrieval scope allows the model to access sufficient diverse temporal patterns for effective generalization. However, a further increase in $\kappa$ (to $0.9$ or $1.0$) results in a performance decline. This phenomenon verifies our hypothesis that the knowledge bank contains inevitable semantic noise or task-irrelevant temporal dynamics.
Therefore, $\kappa=0.7$ strikes an optimal balance between informativeness and robustness.
\begin{table}[t]
\centering
\caption{Impact of momentum coefficient $\mu$ in IKT under the $5$-way $1$-shot setting on SSv2-small~\citep{goyal2017something} and HMDB51~\citep{kuehne2011hmdb}~(see \S\ref{ex::abs}). The adopted design is marked in \textcolor{red}{red}.}
\vspace{-5pt}
\label{tab::momentum}
\resizebox{\columnwidth}{!}{
\setlength{\tabcolsep}{10pt}    
\renewcommand\arraystretch{1.0}
\begin{tabular}{c|cc}
\thickhline
\rowcolor{mygray}
 Momentum Coefficient $\mu$ & SSv2-small & HMDB51  \\ \hline  \hline
  $\mu=0$         & $57.2$    & $83.1$     \\  
       $\mu=0.5$       & $59.0$   & $84.4$     \\ 
   $\mu=0.9$      & $59.4$       & $85.0$     \\ 
\textcolor{red}{$\mu=0.99$}      & $\mathbf{60.2}$    & $\mathbf{85.6}$       \\ 
 $\mu=0.999$      & $59.9$        & $85.4$    \\       \hline
\end{tabular}}\vspace{-12pt}
\end{table}

\subsubsection{Impact of Temporal Prototype Number $M$ in IKT}
\label{sec:prototype}
Table~\ref{tab::prototype} reports the performance of our HR$^{2}$G-shot with regard to the number of temporal prototypes, which determines the granularity of the summarized temporal representations. The case of $M=1$ means that directly treating one video as a global representation. 
The suboptimal performance ($59.0\%$ on SSv2-small~\citep{goyal2017something}) suggests that global pooling overlooks critical fine-grained temporal cues for action recognition.
Our algorithm performs best with $M=3$ temporal prototypes.
We hypothesize that $M=3$ aligns well with the intrinsic structure of actions (typically consisting of onset, apex, and offset phases), thus effectively summarizing discriminative patterns.
However, further increasing $M$ to $8$ causes a performance drop. This indicates that maintaining excessive temporal granularity (similar to raw frame-level features) introduces redundancy and noise, hampering the generalization in the few-shot regime.

\subsubsection{Impact of Momentum Coefficient $\mu$ in IKT}
\label{sec:momentum}
We probe the impact of momentum coefficient $\mu$ (Eq.~\ref{eq:miu}) on SSv2-small~\citep{goyal2017something} and HMDB51~\citep{kuehne2011hmdb}, which regulates the update rate of the memory representations.  The results on SSv2-small~\citep{goyal2017something} and HMDB51~\citep{kuehne2011hmdb} are summarized in Table~\ref{tab::momentum}.
We can clearly observe that, the performance of our algorithm improves significantly as $\mu$ increases, peaking at $\mu=0.99$ ($60.2\%$ and $85.6\%$). 
This suggests that a slow and smooth update strategy is essential for maintaining stable and consistent memory representations, effectively mitigating feature fluctuations from individual batches.
However, when $\mu$ approaches $1.0$ (\emph{e.g.}, $\mu=0.999$), the performance drops slightly. We attribute this to the memory becoming overly inert, failing to adapt timely to the evolving feature space during training. Notably, at the extreme of $\mu=0$ (\emph{i.e.}, no momentum), the performance degrades sharply ($57.2\%$). This confirms that without the stabilization provided by momentum, the memory bank suffers from rapid changes and noise, hindering effective knowledge transfer.
\begin{figure}[!t]
\centering
\includegraphics[width=0.49\textwidth]{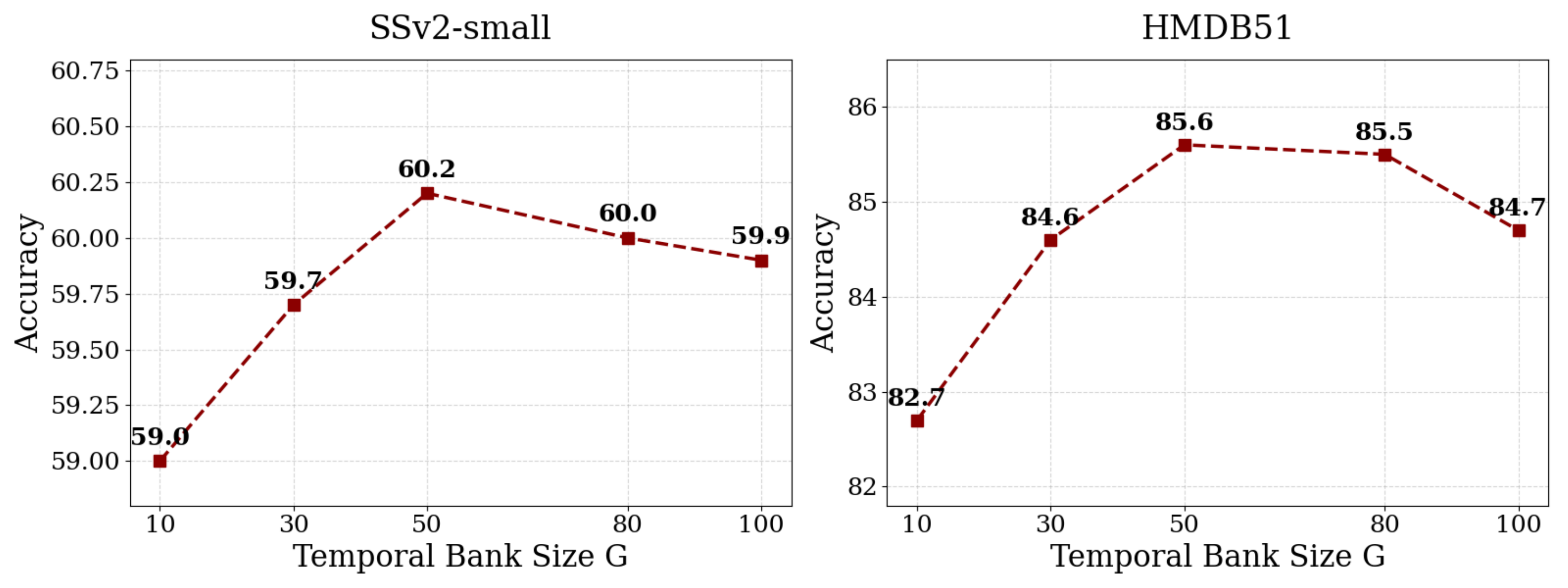}
\vspace{-15pt}
    \caption{The impact of temporal bank size $G$ in IKT on SSv2-small~\citep{goyal2017something} and HMDB51~\citep{kuehne2011hmdb} under the $5$-way $1$-shot setting (\S\ref{ex::abs}).}
    \vspace{-3pt}
    \label{fig::banksize}
\end{figure}

\begin{table}[t]
\caption{Impact of module cascading order under the $5$-way $1$-shot setting on SSv2-small~\citep{goyal2017something} and HMDB51~\citep{kuehne2011hmdb}~(see \S\ref{ex::abs}).}
\vspace{-5pt}
\label{tab:module_order}
\centering
\resizebox{0.44\textwidth}{!}{ 
\setlength\tabcolsep{10pt}
\renewcommand\arraystretch{1}
\begin{tabular}{c|cc}
\thickhline
\rowcolor{mygray}
Arrangement & HMDB51 & SSv2-small \\ \hline \hline
IKT $\rightarrow$ ISC & $83.8$ & $59.1$ \\
\textbf{ISC $\rightarrow$ IKT (Ours)} & $\mathbf{85.6}$ & $\mathbf{60.2}$ \\ \hline
\end{tabular}} 
\vspace{-12pt}
\end{table}

\subsubsection{Impact of Bank Size $G$ in IKT} 
\label{sec:banksize}
Then we study the influence of our knowledge bank on SSv2-small~\citep{goyal2017something} and HMDB51~\citep{kuehne2011hmdb} in Fig.~\ref{fig::banksize}, which determines the capacity for storing representative temporal representations. As seen from Fig.~\ref{fig::banksize}, our algorithm HR$^{2}$G-shot gains stable improvements (\emph{i.e.}, $59.0\%$$\rightarrow$$60.2\%$) on SSv2-small~\citep{goyal2017something} as the bank size grows (\emph{i.e.}, $G=50$). This demonstrates that \textbf{i)} there indeed exist some task-shared temporal patterns in the bank, and  \textbf{ii)} making use of these task-shared temporal patterns can ensure a sufficient temporal understanding for each task. 
However, extending the size beyond $50$ (\emph{e.g.}, $G=80$ or $100$) leads to performance saturation or even a slight decline. We attribute this to the inclusion of redundant or less relevant patterns, which may act as retrieval noise during the retrieval process. Thus, we set bank size $G$ to $50$  to achieve a better trade-off between accuracy and computation costs.

\subsubsection{Module Arrangement Strategy}
\label{sec:module_order}
To validate our ``coarse-to-fine'' design philosophy, we evaluate the impact of the cascading order between Inter-video Semantic Correlation (ISC) and Inter-task Knowledge Transfer (IKT). 
As shown in Table~\ref{tab:module_order}, the default coarse-to-fine order (ISC $\rightarrow$ IKT) consistently yields superior performance, surpassing the reverse order by 1.8\% and  1.1\% on HMDB51 and SSv2-small, respectively.
This performance degradation in the reversed order suggests that performing IKT on uncalibrated features may introduce noise or misalignment. 
These findings confirm that inter-video relation modeling is essential before conducting reliable cross-task knowledge generalization.

\subsubsection{Efficiency Analysis} 
\label{sec:efficiency}
To analyze the effectiveness of our HR$^{2}$G-shot, we list comparison results with CLIP-FSAR~\citep{wang2024clip} in terms of trainable parameters, GPU memory, and inference speed in Table~\ref{efficiency}. We choose ViT-B as our visual encoder. Note that different from the full model parameters in Table~\ref{core_module}, Table~\ref{efficiency} reports trainable parameter comparison. 
As shown in Table~\ref{efficiency}, our HR$^{2}$G-shot introduces a lightweight design with only 19.9M trainable parameters, which is substantially lower than the 89.5M used by CLIP-FSAR.
Although our approach incurs a marginal overhead in inference time ($3.2$ms delay) and memory consumption ($0.1$G increase) due to the hierarchical reasoning modules, it yields substantial performance gains. Specifically, we achieve improvements of 8.5\% and 5.6\% in classification accuracy on HMDB51~\citep{kuehne2011hmdb} and SSv2-small~\citep{goyal2017something}, respectively. These results validate the effectiveness of our design, confirming that the additional computational cost translates directly into better representation capability.
\begin{figure}[!t]
    \centering \includegraphics[width=0.49\textwidth]{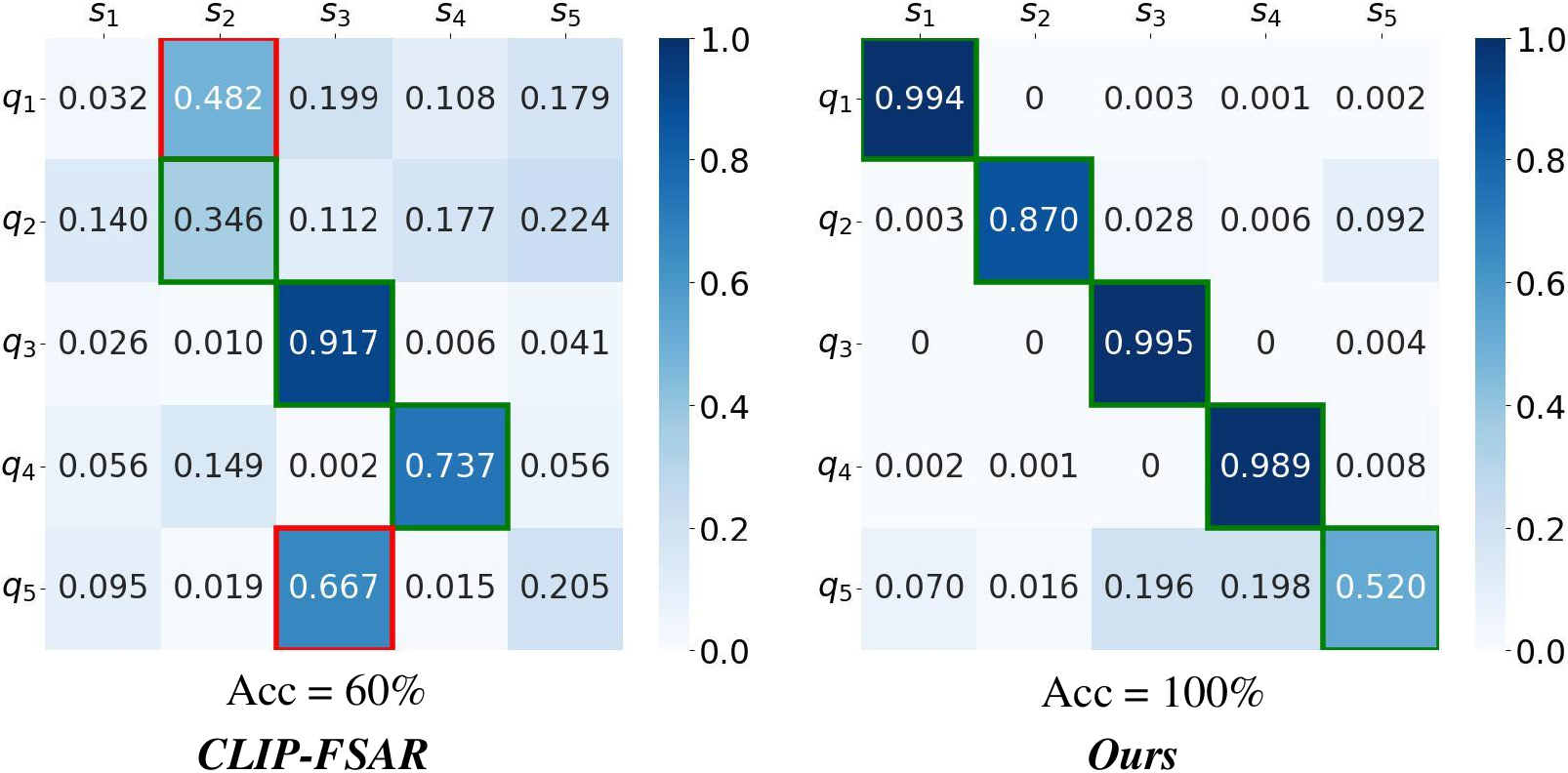}
    \vspace{-15pt}
    \caption{Similarity visualization between query samples ($q_{n}$) and support prototypes ($s_{n}$) with different methods in a meta-test episode from HMDB51~\citep{kuehne2011hmdb} (see \S\ref{sec::quality}). A higher score indicates a greater degree of similarity. The {\color{green} green} box indicates correct prediction and the {\color{red} red} box indicates incorrect prediction.}
    \label{fig::simi}
    \vspace{-2pt}
\end{figure}

\begin{table}[t]
 \caption{Complexity analysis for $5$-way $1$-shot HMDB51~\citep{kuehne2011hmdb} and SSv2-small~\citep{goyal2017something} evaluation. Note that there is $1$ query sample per class. We report trainable parameters, GPU memory, and speed for each model. ``Acc$^{1}$'' and ``Acc$^{2}$'' are the accuracy on HMDB51 and SSv2-small, respectively (\S\ref{ex::abs}). }
 \vspace{-5pt}
    \label{efficiency}
\centering
\resizebox{\columnwidth}{!}{
    \setlength\tabcolsep{3pt}
    \renewcommand\arraystretch{1.1}
    \begin{tabular}{c|c|c|c|c|c}
\thickhline
\rowcolor{mygray} Method  & Params  & Memory & Speed & Acc$^{1}$ &Acc$^{2}$ \\
 \hline\hline
 CLIP-FSAR          & $89.5$M  &$14.2$G&$36.7$ms & $77.1$ & $54.6$  \\ 
 \textbf{HR$^{2}$G-shot (Ours)}   & $\mathbf{19.9}$M  &$\mathbf{14.3}$G &$\mathbf{39.9}$ms  &  $\mathbf{85.6}$  & $\mathbf{60.2}$   \\ \hline 
\end{tabular}}
\vspace{-12pt}
\end{table}

\subsection{Qualitative Analysis}
\label{sec::quality}
\subsubsection{Similarity Visualization} 
\label{sec:simi}
To qualitatively demonstrate the effectiveness of hierarchical relation modeling in our HR$^{2}$G-shot, we visualize the predicted similarities between query and support prototypes with different approaches (\emph{i.e.}, CLIP-FSAR~\citep{wang2024clip} and ours) for one task of HMDB51 in Fig.~\ref{fig::simi}. As seen, our HR$^{2}$G-shot can make more accurate decisions for similar classes in each task compared with CLIP-FSAR. Specifically, for the first query sample in Fig.~\ref{fig::simi}, the incorrect decision obtained by CLIP-FSAR can be rectified by jointly modeling three types of relations. These results further demonstrate that our method can  effectively capture task-specific temporal cues by transferring knowledge at both the video level and the task level.
\begin{figure*}[!t]
    \centering
\includegraphics[width=0.97\textwidth]{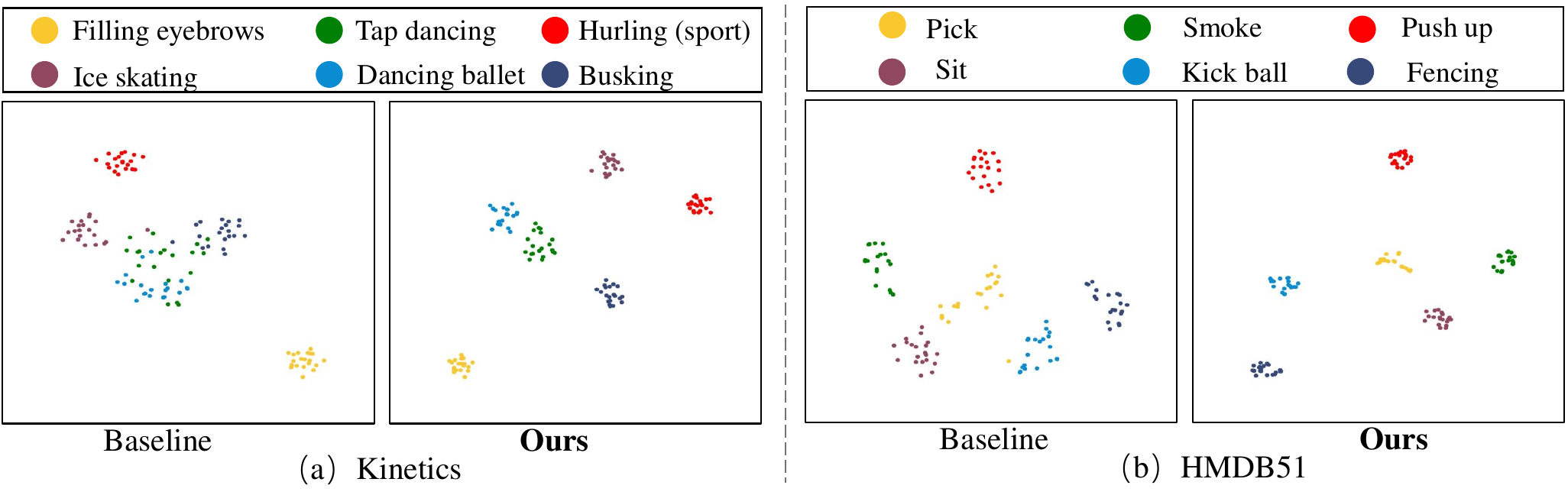}
   \vspace{-15pt}
    \caption{T-SNE feature visualization  of six classes learned by baseline and HR$^{2}$G-shot on Kinetics~\citep{carreira2017quo} and HMDB51~\citep{kuehne2011hmdb} (\S\ref{sec::quality}).}
    \vspace{-10pt}
    \label{fig::tSNE}
\end{figure*}

\subsubsection{The Visualization of Feature Distribution} 
\label{sec:tSNE}
Fig.~\ref{fig::tSNE} visualizes learned features of baseline and our algorithm HR$^{2}$G-shot with ISC and IKT via the t-SNE tool. We clearly observe that after conducting inter-video and inter-task interactions, learned video features become more compact (intra-class) and better separated (inter-class).  As shown in Fig.~\ref{fig::tSNE}(b), the three classes ``Smoke", ``Sit", and ``Pick" become
clearly distinguishable from each other after conducting hierarchical relation modeling. The above phenomena suggest that we can gain task-specific temporal features by transferring knowledge at both video-level and task-level, ensuring better feature discrimination.

\subsubsection{Attention Visualization of Our HR$^{2}$G-shot} 
\label{sec:attn}
\begin{figure}[!t]
    \centering
        \includegraphics[width=0.5\textwidth]{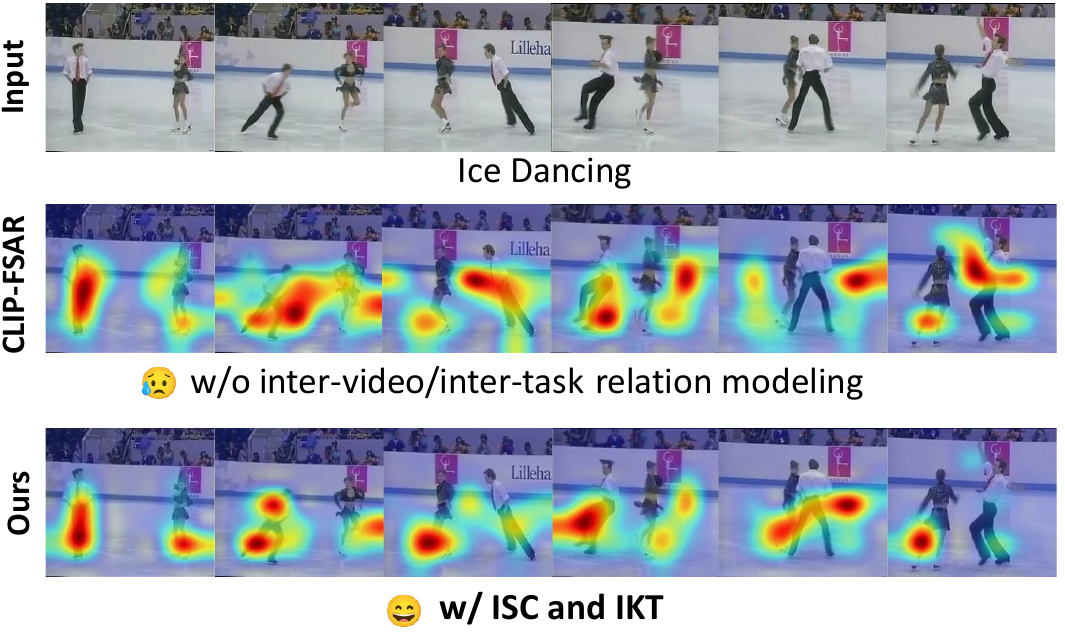}
         \vspace{-15pt}
		\caption{Attention visualization of our HR$^{2}$G-shot on Kinetics in the 5-way 1-shot setting (\S\ref{sec::quality}). Corresponding to the original RGB images, the attention maps of CLIP-FSAR~\citep{wang2024clip} without inter-video and inter-task relation modeling are compared to the attention maps produced by our HR$^{2}$G-shot.}
		\label{fig::attn}	
        \vspace{-4.3mm}
	\end{figure}

To qualitatively evaluate the effectiveness of the proposed HR$^{2}$G-shot, we visualize the attention maps on Kinetics~\citep{carreira2017quo} under the 5-way 1-shot setting, as illustrated in Fig.~\ref{fig::attn}. We compare our approach with the baseline CLIP-FSAR~\citep{wang2024clip}, which lacks explicit inter-video and inter-task relation modeling. The baseline attention maps are often fragmented—distracted by unrelated background objects, while our HR$^{2}$G-shot generates more concentrated and precise activations. By integrating the Inter-video Semantic Correlation (ISC) and Inter-task Knowledge Transfer (IKT) modules, our model effectively suppresses task-irrelevant noise and prioritizes action-salient regions. For example, in the ``Ice Dancing" sequence, our model consistently anchors its focus on the rhythmic movements of the dancers rather than the static rink environment, thereby capturing the key motion cues. These attention visualization maps provide empirical evidence that hierarchical relation modeling enables the extraction of more discriminative, task-specific temporal features.
\begin{table}[t]
\caption{Generalization performance of our method with different temporal alignment metrics on  SSv2-small~\citep{goyal2017something} and HMDB51~\citep{kuehne2011hmdb} under the $5$-way $1$-shot setting (\S\ref{sec::gs}).}
\vspace{-7pt}
\label{table6}
\centering
\resizebox{\columnwidth}{!}{
 \setlength\tabcolsep{9pt}
    \renewcommand\arraystretch{1.1}
  \begin{tabular}{l|cc}
\thickhline
\rowcolor{mygray} Temporal Metric  & SSv2-small & HMDB51   \\ \hline \hline
  
   CLIP-FSAR (Bi-MHM)      & $54.1$ &  $77.0$  \\ 
  \textbf{Ours (Bi-MHM)}      & $\mathbf{60.1}$  & $\mathbf{85.0}$   \\  \hline
  CLIP-FSAR (TRX)       & $53.8$ &  $77.4$  \\ 
      \textbf{Ours (TRX)}  & $\mathbf{60.4}$   & $\mathbf{85.7}$   \\ \hline

   CLIP-FSAR (OTAM)       & $54.6$ &  $77.1$  \\ 
      \textbf{Ours (OTAM)}  & $\mathbf{60.2}$   & $\mathbf{85.6}$   \\ \hline
\end{tabular}}
\vspace{-0.3cm}
\end{table}

\begin{table}
\caption{Comparison with SOTAs using the ResNet-based CLIP backbone under the 5-way 1-shot setting (\S\ref{sec::gs}).}
\label{tab:rn50_comp}
 \vspace{-6pt}
\centering
\small
\resizebox{\columnwidth}{!}{ 
\setlength\tabcolsep{5pt} 
\renewcommand\arraystretch{0.95}
\begin{tabular}{c|ccc}
\thickhline
\rowcolor{mygray}
Method & SSv2-small & UCF101 & HMDB51 \\ \hline \hline
CLIP-FSAR~\cite{wang2024clip} & $52.0$ & $91.3$ & $72.5$ \\
MVP-shot~\cite{qu2024mvp} & $51.2$ & $92.2$ & $72.5$ \\ \hline
\textbf{HR$^{2}$G-shot (ours)}  & $\mathbf{53.1}$ & $\mathbf{93.7}$ & $\mathbf{74.5}$ \\ \hline
\end{tabular}} 
 \vspace{-6pt}
\end{table}

	\begin{figure*}[!t]
    \centering
        \includegraphics[width=0.90\textwidth]{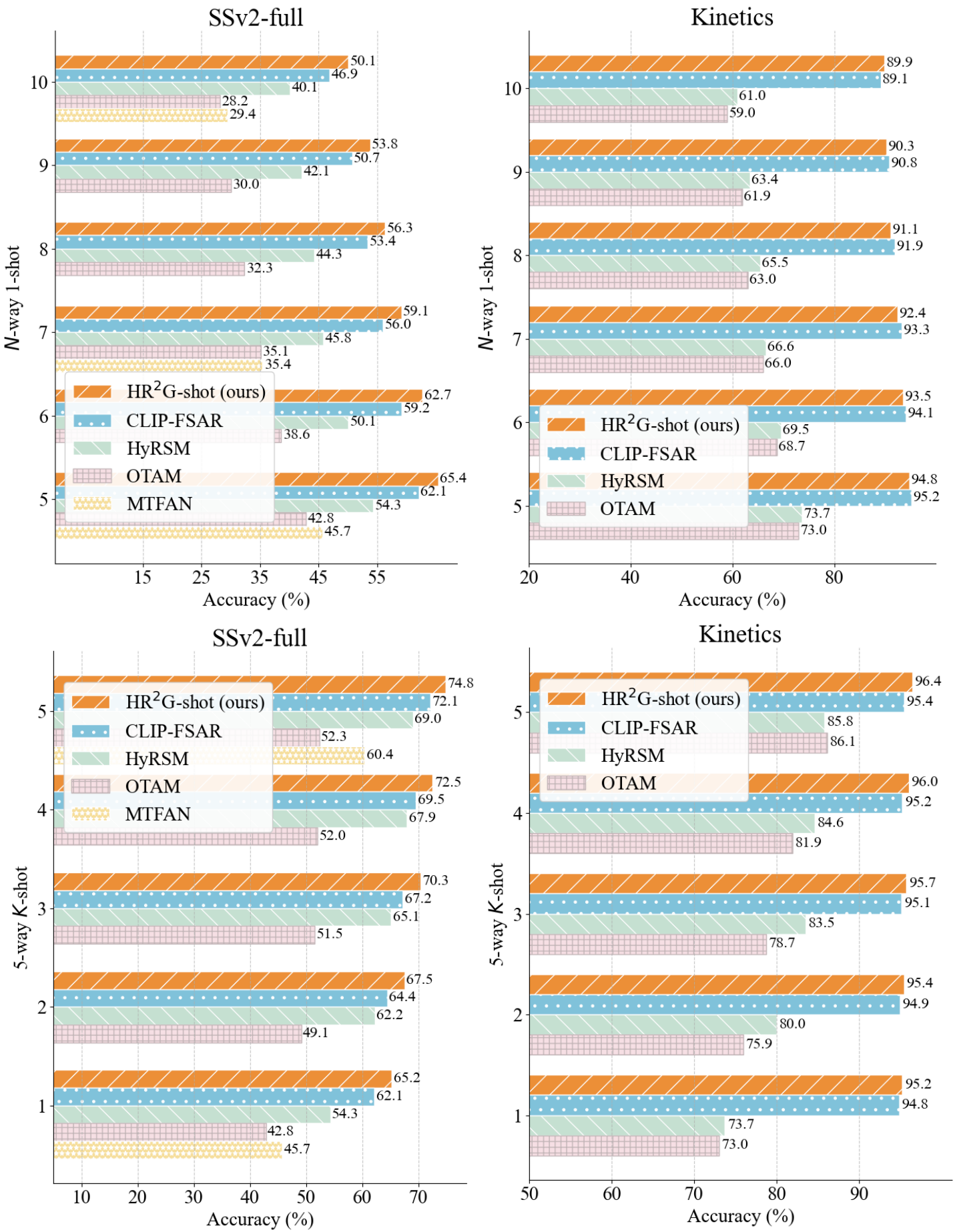}
        \vspace{-10pt}
		\caption{Generalization performance of our method with different ways and shots on SSv2-full~\citep{goyal2017something} and Kinetics~\citep{carreira2017quo} (\S\ref{sec::gs}).}
		\label{fig::nway}	
        \vspace{-8pt}
	\end{figure*}

\subsection{Generalization Study}
\label{sec::gs}
\subsubsection{Results with Various Metrics} 
\label{sec:table6}
To demonstrate that HR$^{2}$G-shot generalizes well to different alignment metrics, Table~\ref{table6} reports comparison results on HMDB51~\citep{kuehne2011hmdb} and SSv2-small~\citep{goyal2017something} with regard to different temporal metrics (\emph{cf.} Eq.~\ref{eq::temporal_metric}). Different temporal metrics are adopted, including OTAM~\citep{cao2020few}, TRX~\citep{perrett2021temporal} and Bi-MHM~\citep{wang2022hybrid}. 
The results clearly indicate that the benefits of our HR$^{2}$G-shot are not limited to a specific alignment strategy. Our method consistently outperforms CLIP-FSAR by significant margins across all settings. Notably, while the baseline performance fluctuates depending on the metric, our method maintains high accuracy (\emph{e.g.}, $\sim$$60\%$ on SSv2-small and $\sim$$85\%$ on HMDB51). This demonstrates that the hierarchical relation modeling in our framework learns intrinsically better video representations, thereby unlocking the potential of downstream alignment metrics.
\subsubsection{Influence of Different Visual Backbones}
\label{sec:rn50_comp}
Here, we perform a comprehensive evaluation to investigate the generalization capability of our framework across different architectural designs. 
Specifically, we adopt the ResNet-based CLIP backbone as the visual encoder, while keeping all other algorithmic configurations identical to our default setting. 
The comparative results against state-of-the-art methods, including CLIP-FSAR~\citep{wang2024clip} and MVP-shot~\citep{qu2024mvp}, are reported in Table~\ref{tab:rn50_comp}. From Table~\ref{tab:rn50_comp}, it is evident that our proposed method consistently outperforms existing competitors on the CLIP-RN50 architecture. 
For instance, under the 1-shot setting, our model surpasses the strong baseline MVP-shot by 1.9\%, 1.5\%, and 2.0\% on SSv2-small~\citep{goyal2017something}, UCF101~\citep{soomro2012ucf101}, and HMDB51~\citep{kuehne2011hmdb}, respectively. 
This suggests that our approach is backbone-agnostic and can effectively extract discriminative relational features regardless of the underlying visual encoder.
\subsubsection{Results with Various Ways and Shots} 
\label{sec:nway}
We compare HR$^{2}$G-shot with state-of-the-art methods, including CLIP-FSAR~\citep{wang2024clip}, HyRSM~\citep{wang2022hybrid}, OTAM~\citep{cao2020few}, and MTFAN~\citep{wu2022motion}, across both $N$-way 1-shot and 5-way $K$-shot scenarios. As shown in Fig.~\ref{fig::nway}, HR$^{2}$G-shot consistently achieves superior performance across various $N$-way and $K$-shot settings on both SSv2-full~\citep{goyal2017something} and Kinetics~\citep{carreira2017quo} datasets. Specifically, on the motion-heavy SSv2-full dataset, our HR$^{2}$G-shot outperforms the runner-up CLIP-FSAR by 3.2\% in the 10-way 1-shot setting (50.1\% vs. 46.9\%) and maintains a substantial gain of 14.1\% over MTFAN in the 5-way 5-shot configuration (74.5\% vs. 60.4\%). On Kinetics, HR$^{2}$G-shot achieves peak accuracy of 96.4\% in the 5-way 5-shot scenario. These results demonstrate that our hierarchical relation-augmented representations effectively enhance generalization across diverse and demanding few-shot action recognition tasks.

\section{Conclusion and Discussion}
\sssection{Conclusion.} In this work, we present HR$^{2}$G-shot, a novel Hierarchical Relation-augmented Representation Generalization framework for FSAR, which captures task-specific temporal cues from multiple perspectives (\emph{i.e.}, inter-frame, inter-video, and inter-task). Rather than only conducting inter-frame temporal interactions, we further \textbf{i)} perform fine-grained cross-video temporal interaction for exploring inter-video relationships, and \textbf{ii)} aggregate useful temporal knowledge from previous tasks for learning inter-task relationships. Our framework is hierarchical, elegant, and gains outstanding performance on five standard FSAR datasets, \emph{i.e.}, SSv2-full, SSv2-small, HMDB51, UCF101, and Kinetics. Our work offers a fresh perspective to current FSAR solutions by hierarchically exploring multi-level relations, and we wish it to inspire future research in this field.

\sssection{Discussion.} One limitation of our algorithm is that it retrieves and aggregates temporal knowledge from historical tasks, yet ignores spatial information in actions. In the future, we plan to upgrade the knowledge bank to jointly encode and transfer task-specific spatio-temporal patterns. Additionally, while our work primarily focuses on few-shot action recognition tasks, the proposed hierarchical relational paradigm possesses high versatility. In the future, we aim to extend this framework to a broader range of applications, including zero-shot action recognition and other related tasks.




{
\bibliographystyle{IEEEtran}
\bibliography{sn-bibliography}
}

\end{document}